\documentclass[10pt,twocolumn,letterpaper]{article}

\usepackage[pagenumbers]{cvpr} % To force page numbers, e.g. for an arXiv version
\usepackage{verbatim}
\usepackage{lipsum}
\usepackage[table]{xcolor}
\usepackage{multirow}

\usepackage{pifont}
\usepackage{newunicodechar}
\newunicodechar{✓}{\ding{51}}
\newunicodechar{✗}{\ding{55}}
\usepackage{graphicx}
\usepackage{subcaption}
\usepackage{xcolor}
\usepackage{tikz}
\usepackage{adjustbox}
\usepackage{array}

\usepackage[accsupp]{axessibility}  % Improves PDF readability for those with disabilities.

\usepackage{xr-hyper}
\newcommand{\methodname}{Dark3R\xspace}

\newcommand{\R}{\mathbb{R}}

\newcommand{\img}[1]{\mathbf{I}^{({#1})}}
\newcommand{\cleanimg}[1]{\mathbf{I}_\text{clean}^{({#1})}}
\newcommand{\noisyimg}[1]{\mathbf{I}_\text{noisy}^{({#1})}}

\newcommand{\enc}{\mathcal{E}}            % encoder
\newcommand{\dec}{\mathcal{D}}            % decoder (twin)
\newcommand{\noisyenc}{\mathcal{\tilde{E}}}            % encoder
\newcommand{\noisydec}{\mathcal{\tilde{D}}}            % decoder (twin)
\newcommand{\head}{\mathcal{H}}          % lightweight heads
\newcommand{\noisyhead}{\mathcal{\tilde{H}}}          % lightweight heads

\newcommand{\fenc}[1]{\mathbf{F}_{\enc}^{(#1)}}

\newcommand{\f}{\mathbf{F}}
\newcommand{\noisyf}{\mathbf{\tilde{F}}}
\newcommand{\noisyfenc}[1]{\mathbf{\tilde{F}}_{\enc}^{(#1)}}

\newcommand{\pointmap}{\mathbf{P}}                 % per-pixel 3D point map
\newcommand{\corrmap}{\mathbf{C}}                  % correspondence / match map
\newcommand{\noisycorrmap}{\mathbf{\tilde{C}}}                  % correspondence / match map

\definecolor{cvprblue}{rgb}{0.21,0.49,0.74}
\definecolor{tabfirst}{rgb}{1, 0.7, 0.7} % red
\definecolor{tabsecond}{rgb}{1, 0.85, 0.7} % orange
\definecolor{tabthird}{rgb}{1, 1, 0.7} % yellow
\usepackage[pagebackref,breaklinks,colorlinks,allcolors=cvprblue]{hyperref}

\def\paperID{*****}\def\confName{***}
\def\confYear{***}

\title{\methodname: Learning Structure from Motion in the Dark}

\author{
    Andrew Y. Guo\textsuperscript{1,2} \quad
    Anagh Malik\textsuperscript{1,2} \quad
    SaiKiran Tedla\textsuperscript{3} \quad
    Yutong Dai\textsuperscript{4} \\
    Yiqian Qin\textsuperscript{1,2} \quad
    Zach Salehe\textsuperscript{5} \quad
    Benjamin Attal\textsuperscript{1} \\
    Sotiris Nousias\textsuperscript{1,6} \quad
    Kiriakos N. Kutulakos\textsuperscript{1,2} \quad
    David B. Lindell\textsuperscript{1,2} 
    \vspace{0.5em} \\ % Larger space before affiliations
    \textsuperscript{1}University of Toronto \quad
    \textsuperscript{2}Vector Institute \quad
    \textsuperscript{3}York University \\
    \textsuperscript{4}Sony Corporation of America \quad
    \textsuperscript{5}Harvard University \quad
    \textsuperscript{6}Purdue University 
    \\[0.4em]
    {\href{https://andrewguo.com/pub/dark3r}{\textcolor{cvprblue}{\texttt{andrewguo.com/pub/dark3r}}}}
    \\[-1.8em]
}
\begin{document}

% \maketitle

\twocolumn[{
    \maketitle
    \begin{center}
    \includegraphics[width=\textwidth]{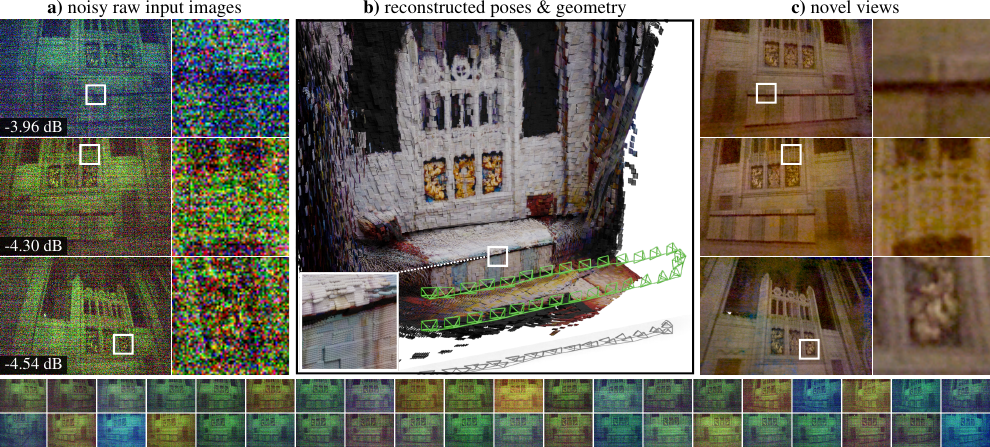}
    \end{center}
    \vspace{-1em}
    \captionof{figure}{
\methodname enables structure from motion and novel view synthesis from raw images captured in low-light conditions.
\textbf{(a)} For this scene, we capture 500 images from varying viewpoints and show a subset along with their signal-to-noise ratios. Temporal sensor noise causes pronounced frame-to-frame color variations, visible in the bottom row, which further complicates the problem.
\textbf{(b)} We apply \methodname to these images to recover camera poses and the 3D scene geometry (we show a subset of the predicted poses).
\textbf{(c)} Finally, we introduce a robust view-synthesis technique that leverages \methodname{’s} predicted poses and a coarse-to-fine optimization strategy to reconstruct fine appearance details that are otherwise completely obscured by noise. \textbf{Please refer to the project webpage for video results.}}
    \label{fig:teaser}
    \vspace{1em}
}]

% \makeatletter
% \renewenvironment{abstract}{
%       \section*{{\vspace{-2em}
% \begin{center}\abstractname\end{center}}\vspace{-0.75em}}\itshape%
% }{\par}
% \makeatother

\begin{abstract}
% \vspace{-0.5em}
We introduce \methodname, a framework for structure from motion in the dark that operates directly on raw images with signal-to-noise ratios (SNRs) below $-4$ dB—a regime where conventional feature- and learning-based methods break down.
Our key insight is to adapt large-scale 3D foundation models to extreme low-light conditions through a teacher–student distillation process, enabling robust feature matching and camera pose estimation in low light.
\methodname requires no 3D supervision; it is trained solely on noisy--clean raw image pairs, which can be either captured directly or synthesized using a simple Poisson–Gaussian noise model applied to well-exposed raw images.
To train and evaluate our approach, we introduce a new, exposure-bracketed dataset that includes $\sim$42,000 multi-view raw images with accurate 3D annotations, and we demonstrate that \methodname\ achieves state-of-the-art structure from motion in the low-SNR regime. 
Further, we demonstrate state-of-the-art novel view synthesis in the dark using \methodname's predicted poses and a coarse-to-fine radiance field optimization procedure. 

\vspace{-1.5em}
\end{abstract}

% Passive 3D reconstruction methods such as structure-from-motion (SfM) and stereo vision have achieved remarkable success under well-lit conditions but fail catastrophically in extreme low light, where sensor noise overwhelms image features.
    
\begin{figure*}[ht]
\includegraphics[width=\textwidth]{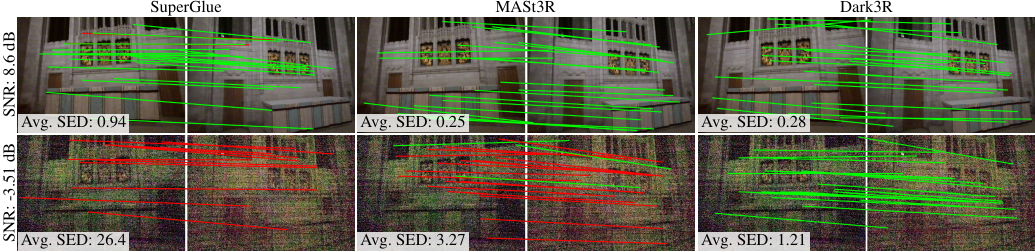}
\vspace{-1.5em}
\caption{Existing hand-crafted and data-driven feature-matching pipelines such as SuperGlue~\cite{sarlin2020superglue} and MASt3R~\cite{leroy2024grounding} perform reliably under well-illuminated conditions (top row) but performance significantly worsens when the image signal-to-noise ratio (SNR) drops to below –3 dB (bottom row). In contrast, \methodname{} robustly identifies corresponding points in both imaging regimes. Green and red lines denote correspondences whose symmetric epipolar distance (SED) is below or above two pixels, respectively, for a randomly selected set of 20 putative matches. The average SED over all matches is also reported. We compute SED using calibrated camera intrinsics and the essential matrix predicted from MASt3R correspondences on the high-SNR image pair.} 
\vspace{-1.5em}
\label{fig:snr}
\end{figure*}

\section{Introduction}
\label{sec:intro}

Passive techniques for 3D reconstruction, such as stereo vision and structure from motion (SfM), have been developed over decades \cite{scharstein2002taxonomy,hartley2003multiple,szeliski2010computer,schonberger2016structure} and underpin modern frameworks for reconstructing appearance and geometry from captured or generated images \cite{mildenhall2021nerf,xu2024grm}.
However, despite their maturity and widespread success, passive 3D reconstruction methods still fail under low-light conditions, where noise dominates the captured signal.
We seek to address this problem by enabling robust SfM in extreme low-light environments---or with image signal-to-noise ratios (SNR) well below 0 dB (see Fig.~\ref{fig:teaser}).

Conventional SfM methods recover camera poses and scene geometry jointly through a multi-stage pipeline that involves detecting and matching image features, estimating epipolar geometry, performing triangulation, and refining the solution with bundle adjustment \cite{hartley2003multiple,schonberger2016structure}.
In recent years this pipeline has been substantially improved—for example, learning-based feature detection \cite{detone2018superpoint} and matching \cite{sarlin2020superglue} now outperforms conventional hand-crafted techniques \cite{lowe2004distinctive}, and differentiable RANSAC \cite{brachmann2017dsac,brachmann2019dsacpp} enables more robust camera pose estimation in the presence of outliers.
However, despite these advances, the SfM pipeline collapses in low-light conditions because noise causes existing feature detection and matching techniques to fail (see Fig.~\ref{fig:snr}).

More recent methods seek to replace the SfM pipeline with foundation models based on vision transformers and large-scale training datasets \cite{leroy2024grounding,wang2024dust3r,wang2025vggt} or end-to-end optimized neural volumetric scene representations \cite{lin2021barf, yang2024colmapfree, tang2024gaussian}.
Although these methods often achieve higher reconstruction quality than conventional SfM pipelines, they struggle to generalize to low-light conditions because low-SNR images create spurious local minima or deviate from their training distribution (e.g., due to sensor-specific noise statistics).

The fundamental obstacle to applying existing SfM techniques to low light is that their feature extraction modules, whether hand-designed or learned, fail in the presence of significant noise.
This breakdown propagates to downstream components such as camera pose estimation and triangulation.
One possible solution to low-light SfM is to increase image exposure time; but without a tripod-mounted capture setup, hand shake can result in significant motion blur \cite{delbracio2015burst,su2017deep}.
Another option is to apply off-the-shelf denoisers \cite{zhang2017dncnn,brooks2019unprocessing} to the noisy low-light images and feed the result into an SfM method.
However, this naive solution yields inaccurate results because it does not preserve multi-view consistent image features (see Supp.\ Fig.~\ref*{fig:supp-denoising-feature-matching}).

Here, we introduce \methodname, an end-to-end framework for SfM in the dark. 
Our key insight is to adapt the strong priors learned by recent 3D foundation models such as MASt3R~\cite{leroy2024grounding,duisterhof2025mast3r} to the low-light regime. 
Inspired by teacher–student knowledge distillation~\cite{bucilua2006model,hinton2015distilling,zhang2019your,gou2021knowledge}, we develop a training strategy that aligns the dense feature maps predicted by MASt3R on well-exposed raw image pairs with those produced by a student model on low-light raw image pairs.
Crucially, \methodname requires no 3D supervision; it is trained solely on noisy–clean raw image pairs, which can be either captured directly or synthesized using a simple Poisson–Gaussian noise model applied to well-exposed raw images.
After training \methodname, we use correspondences from the predicted feature maps and follow the global optimization and bundle adjustment stages of MASt3R-SfM~\cite{duisterhof2025mast3r} to recover camera poses and sparse depth maps from multi-view, noisy raw images.

To train and evaluate \methodname, we introduce a new, first-of-its kind dataset of roughly 42,000 multi-view, exposure-bracketed raw images with accurate 3D annotations, and a further set of roughly 20,000 high-SNR multi-view raw images across nearly 100 different scenes. 
For evaluation, we provide reference 3D annotations derived from the high-SNR exposures, which serve as a reference for assessing pose accuracy.
Using this dataset, we demonstrate that \methodname enables accurate camera pose and depth estimation where previous methods fail---hence, our work opens up the low-light regime for new applications of passive 3D sensing. 
Moreover, by combining \methodname{’s} predicted poses and depths with a coarse-to-fine radiance field optimization scheme, we enable new capabilities of novel view synthesis in low-SNR settings.
\vspace{-0.3em}

\section{Related Work}
\vspace{-0.3em}

\label{sec:related_work}

\paragraph{Low-light image processing.}
A common approach to improving the quality of images captured in low-light conditions is to denoise each frame independently using state-of-the-art classical or neural denoisers \cite{dabov2007bm3d,zhang2017dncnn,brooks2019unprocessing,yue2019vdn,zhang2019ridnet,chen2018learning, jin2023led}.
While this may improve signal-to-noise ratio (SNR) on a per-image basis, it comes with a key drawback in the context of 3D reconstruction: denoising frames independently fails to enforce multi-view consistency~\cite{pearl2022nan,mildenhall2022rawnerf}.
This can result in incorrect correspondences and, consequently, inaccurate or failed pose estimation.
In contrast, our method learns to find correspondences directly on noisy raw images and denoise them in a multi-view-consistent manner.

Another line of work explores burst denoising \cite{liba2019handheld,mildenhall2018burst,bhat2021burstsr,wang2023burstnerf,ravendran2021burst,chugunov2023shakes,chugunov2024neural}, where a sequence of images captured in rapid succession is jointly processed to improve image quality.
However, burst techniques typically assume small motion between frames---an assumption that breaks down in 3D reconstruction pipelines with moving cameras, where frame-to-frame changes involve substantial parallax and motion blur.
In such cases, burst denoising algorithms fail to align the frames accurately, resulting in artifacts or poor denoising performance.
Instead, our method processes noisy image sequences with significant motion by directly estimating camera poses and enforcing multi-view consistency in reconstructed scene appearance.

\vspace{-1em}
\paragraph{Neural 3D reconstruction.}
Recent works have explored 3D reconstruction in low light by building a multi-view-consistent representation of the scene using raw images or noisy input images \cite{mildenhall2022rawnerf,huang2022hdr,pearl2022nan}.
These methods demonstrate improved performance over 2D single-frame and burst denoising as they integrate inter-view and spatial information from multiple images with larger baselines.
However, they require camera poses from off-the-shelf SfM pipelines and, as a result, do not work below a certain light level.
For example, RawNeRF~\cite{mildenhall2022rawnerf} was demonstrated only down to image SNRs for which COLMAP~\cite{schonberger2016structure} could still recover accurate camera poses.
Our method also operates directly on raw images, but recovers poses and 3D geometry at image SNRs below the threshold where conventional SfM methods fail (see Fig.~\ref{fig:snr}).

\vspace{-1em}
\paragraph{Direct and learned SfM.}
Feature-based pipelines for SfM~\cite{snavely2006photo,schonberger2016structure} detect and match a sparse set of hand-crafted~\cite{lowe2004distinctive,bay2008surf} or learned  keypoints~\cite{detone2018superpoint,sarlin2020superglue} across frames to estimate epipolar geometry and triangulate scene points.
Typically, a final bundle-adjustment step refines camera poses and points by minimizing a geometric objective, most commonly the reprojection error \cite{triggs2000bundle}.
While feature matching excels in well-lit textured scenes, it degrades in low-light settings, causing epipolar geometry estimation to fail.

Alternatively, direct SfM methods operate on image intensities and optimize a photometric consistency cost without explicit keypoints~\cite{newcombe2011dtam,engel2014lsdslam,engel2018dso,zhou2020d3vo,zou2020dfvo,sucar2021imap}.
When inter-frame motion is small, direct methods typically outperform feature-based ones because they leverage all pixel intensities to estimate a few motion parameters.
In low light, however, read noise dominates, violating brightness constancy and causing direct methods to break down.

More recent data-driven approaches learn to find correspondences using neural networks~\cite{sarlin2020superglue,leroy2024grounding,truong2023pdc,duisterhof2025mast3r,zhang2025monst3r}, develop end-to-end differentiable SfM pipelines~\cite{brachmann2019dsacpp}, or directly regress camera poses and 3D structure in a feed-forward manner~\cite{kendall2015posenet,wang2024dust3r,elflein2025light3r,wang2025vggt,keetha2025mapanything}.
These methods often depend on large-scale pretrained models such as DINO~\cite{oquab2023dinov2} or monocular depth estimators~\cite{zhang2022structure,li2025megasam} to pre-process input images.
However, they usually fail to generalize to low-SNR raw input images that lie outside the distribution of their training data.
\methodname overcomes these limitations by learning to detect and match features directly on noisy raw images, enabling robust pose and structure estimation, even for very low SNRs.

\begin{figure*}
\includegraphics[width=\textwidth]
{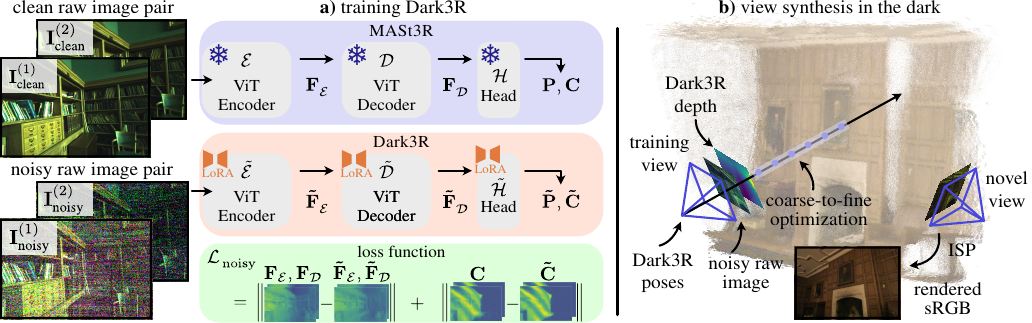}
% {figures/DRAFT_DARK3R.pdf}
\vspace{-2em}
\caption{
    Method overview. \textbf{(a)} \methodname is trained using paired clean and noisy raw images, $(\cleanimg{1}, \cleanimg{2})$ and $(\noisyimg{1}, \noisyimg{2})$.
The model is initialized from the weights of a pretrained MASt3R~\cite{leroy2024grounding} network and adapted to low-light conditions using low-rank adaptation~\cite{hu2022lora}.
We fine-tune the encoder $\noisyenc$,  decoder $\noisydec$, and output head $\noisyhead$. We supervise training by minimizing the difference between MASt3R’s encoder features $\f_\enc$, decoder features $\f_\dec$, and correspondence map $\corrmap$ from the clean pair and \methodname’s predictions $\noisyf_\enc$, $\noisyf_\dec$, and $\noisycorrmap$ on the noisy pair.
\textbf{(b)} After training, the predicted poses and depth maps from \methodname enable view synthesis in the dark via a coarse-to-fine optimization process.
The rendered novel views are passed through an image signal processor (ISP) to produce the final sRGB outputs.
}
\vspace{-1em}
\label{fig:method}
\end{figure*}

\subsection{Background: MASt3R}
\label{sec:background}
As we build on the MASt3R~\cite{leroy2024grounding} and MASt3R-SfM~\cite{duisterhof2025mast3r} architectures to enable SfM in the dark, we provide a brief overview of both of these methods.  

\vspace{-1em}
\paragraph{MASt3R.} Given two input images $\img{1}, \img{2} \in \mathbb{R}^{H \times W \times 3}$, MASt3R~\cite{leroy2024grounding} predicts pixel-wise correspondences and local 3D point maps along with corresponding confidence maps. 
These quantities are predicted by first using a vision transformer (ViT) encoder $\enc$ to predict features $\f_\enc = [\fenc{1}, \fenc{2}] \in\R^{2\times H\times W \times D_\enc}$ for each image individually as 
\begin{equation}
    \f_\enc = [\enc(\img{1}), \enc(\img{2})].
\end{equation}
Then, a Siamese ViT decoder $\dec$ produces a further set of jointly processed feature maps $\f_\dec\in \R^{2 \times H \times W \times D_\dec}$, which share information between the input viewpoints:
\begin{equation}
 \f_\dec = \dec(\fenc{1}, \fenc{2}).
\end{equation}
Finally, the decoder features are processed by separate heads $\head$ to regress, for both input images, per-pixel 3D point maps $\pointmap\in\R^{2\times H \times W \times 3}$ and feature maps $\corrmap\in\R^{2\times H \times W \times D_C}$ used for correspondence matching.
The entire architecture can be summarized as 
\begin{equation}
    \pointmap, \corrmap = \head\left(\dec\left[\enc(\img{1}), \enc(\img{2})\right]\right),
\end{equation}
where we omit the confidence maps for brevity.
  
\vspace{-1em}
\paragraph{MASt3R-SfM.} Given the features \(\fenc{i}\) extracted from each image in a set \(\{\img{i}\}_{i=1}^{I}\), MASt3R-SfM~\cite{duisterhof2025mast3r} builds a co-visibility graph by computing pairwise feature similarities, where edges link image pairs that are likely to observe overlapping regions of the scene.
Then, for each edge in this graph, MASt3R-SfM predicts a per-pixel 3D point map \(\pointmap\) and correspondence map \(\corrmap\) from the associated stereo pair.
A subsequent coarse alignment stage transforms the locally predicted point maps into a common world coordinate frame.
This global reconstruction serves as the initialization for a bundle adjustment step that jointly refines the estimated point maps and camera parameters by minimizing 2D reprojection errors.

\section{Structure from Motion in the Dark}
\label{sec:method}

\methodname uses a student--teacher distillation procedure to adapt MASt3R~\cite{leroy2024grounding} to the low-SNR regime for extracting pointmaps and feature correspondences from image pairs. 
The estimated poses and correspondences are then used for global 3D reconstruction and bundle adjustment following MASt3R-SfM~\cite{duisterhof2025mast3r}. 
A method overview is shown in Fig.~\ref{fig:method}.

\paragraph{Teacher--Student Low-Light Adaptation}
Given a low-SNR image pair $(\noisyimg{1}, \noisyimg{2})$, our goal is to predict accurate pixel correspondences $\corrmap$ and per-pixel 3D pointmaps $\pointmap$. 
In other words, we seek the same output quantities produced by MASt3R~\cite{leroy2024grounding} when applied to a high-SNR image pair $(\cleanimg{1}, \cleanimg{2})$ captured from the same viewpoints. 

To this end, we introduce a teacher--student framework comprising two networks with identical architectures: (1) a frozen teacher $\{\enc, \dec, \head\}$ corresponding to a pre-trained MASt3R model, and (2) a student $\{\noisyenc, \noisydec, \noisyhead\}$ initialized from the same weights, with a trainable encoder $\noisyenc$, decoder $\noisydec$, and output  head $\noisyhead$. 
The student is fine-tuned using low-rank adaptation~\cite{hu2022lora}, and we enforce feature consistency between the teacher and student when processing corresponding high- and low-SNR image pairs.

For both the low- and high-SNR image pairs, we use raw sensor measurements that have been trivially demosaiced by subsampling pixels from each color channel in the Bayer mosaic and averaging the two green channels. 
Using the raw measurements preserves information that would otherwise be lost in conventional image signal processing (ISP) pipelines, which apply non-invertible operations such as black-level subtraction and clipping. 
Empirically, we find that MASt3R performs equally well on these high-SNR demosaiced raw images as on standard sRGB inputs, so we process all the inputs in this raw space.

The teacher--student loss is computed by passing the noisy image pair through the student network as 
\begin{equation}
\noisyhead\left(\noisydec\left[\noisyenc(\noisyimg{1}), \noisyenc(\noisyimg{2})\right]\right),
\end{equation}
which yields intermediate and final outputs $\noisyf_\text{noisy} = [\noisyf_\enc, \noisyf_\dec, \noisycorrmap]$.
The outputs $\f$ of the frozen teacher network are computed in a similar fashion using the clean images. 
Then, we optimize the student to minimize the sum of $L_2$ distances between corresponding encoder, decoder, and correspondence feature maps:
\vspace{-0.3em}
\begin{equation}
 \mathcal{L}_{\text{noisy}} = \rVert\f - \noisyf_\text{noisy}\lVert_2^2.
\end{equation}
\vspace{-0.3em}
To ensure that the student maintains performance across a broad SNR range, we additionally train on clean image pairs passed through the student and teacher networks. 
In this case we compute the loss function
\vspace{-0.3em}
\begin{equation}
 \mathcal{L}_{\text{clean}} = \rVert\f - \noisyf_\text{clean}\lVert_2^2,
\end{equation}
% \vspace{-0.3em}
where $\noisyf_\text{clean}$ denotes intermediate and final output features from applying the student network to a clean image pair. 

The complete training objective is given as
\vspace{-0.3em}
\begin{equation}
    \mathcal{L} = \mathcal{L}_\text{noisy} + \lambda_\text{clean} \mathcal{L}_\text{clean},
    \label{eq:loss}
\end{equation}
\vspace{-0.3em}
where $\lambda_{\text{clean}}$ controls the regularization strength.
\vspace{-1em}
\paragraph{Inference} 
At inference time, we follow the prediction and bundle adjustment procedure introduced in MASt3R-SfM~\cite{duisterhof2025mast3r}.  
That is, given a set of noisy raw images $\{\noisyimg{i}\}_{i=1}^I$, \methodname predicts corresponding features $\{\noisyfenc{i}\}_{i=1}^I$.
We then use these features to construct a scene graph and perform global reconstruction and bundle adjustment as described in Sec.~\ref{sec:background}. 
Our procedure closely follows MASt3R-SfM~\cite{duisterhof2025mast3r}, except that we assume the camera intrinsics are known and so we incorporate an additional regularization term during bundle adjustment to keep the optimized intrinsics close to their calibrated values.

\vspace{-1em}
\paragraph{Implementation details.}
We train our model using the AdamW optimizer with a learning rate of $\eta_0 = 1\times10^{-3}$ and a piecewise exponential decay schedule ($\gamma = 0.99$) updated every $\tfrac{1}{10}$ of an epoch. 
A linear warm-up gradually increases the learning rate from zero to $\eta_0$ over the first half epoch. 
We set {$\lambda_\text{clean}=0.3$}, and the model is trained for up to 15~epochs or until the validation loss plateaus. 
Training is distributed across 8~RTX~A6000 GPUs and completes in approximately 15~hours.
\vspace{-0.5em}
\section{View Synthesis in the Dark} 
\vspace{-0.3em}
Using poses predicted by \methodname enables downstream applications, including low-light view synthesis based on radiance fields.
However, directly optimizing a radiance field using the predicted poses and noisy input images results in relatively poor-quality view synthesis.
Instead, we propose a coarse-to-fine optimization procedure that leverages the point maps from \methodname in addition to the camera poses.  
Please refer to Supp.\ Sec.~\ref*{sec:supp-implementation-details} for implementation details.
Note that we opt to use a NeRF-based representation~\cite{li2022nerfacc} instead of Gaussian splatting~\cite{kerbl2023gsplat}; we found optimizing the discrete placement of Gaussians to be challenging at high noise levels (see comparisons on the project webpage). 

\vspace{-1em}
\paragraph{Coarse-to-fine optimization.} 
We optimize the radiance field by minimizing the difference between predicted and rendered raw images from each viewpoint in a training set.
To avoid overfitting to noise in the images, we use stochastic preconditioning \cite{Ling_2025}---a form of coarse-to-fine optimization that adds Gaussian noise to the location of each ray sample. 
The noise is annealed from a standard deviation of {$1\times10^{-3}$} to zero during the first 30k iterations of optimization, and we continue optimization for 90k iterations. 

\vspace{-1em}
\paragraph{Depth supervision.} 
We also supervise the NeRF using pointmaps predicted by \methodname, following the depth regularization of DS-NeRF~\cite{deng2022depth} with an exponential decay weighting. 
Since the depth maps from \methodname are dense but capture the coarse scene geometry, we gradually downweight the loss as we move from optimizing coarse to fine scene details. 

\vspace{-1em}
\paragraph{Raw input images.} Similar to RawNeRF \cite{mildenhall2022rawnerf}, our method takes raw images as input; however, we do not apply black level subtraction and clipping. 
At low image SNRs, the mean pixel intensity values are very close to the black level, and so clipping would destroy useful information.
Instead, we rely on the aggregation of measurements from multiple views to improve the SNR sufficiently to render clean novel views without clipping. 
\vspace{-0.5em}
\begin{figure*}[ht]
\includegraphics[width=\textwidth]{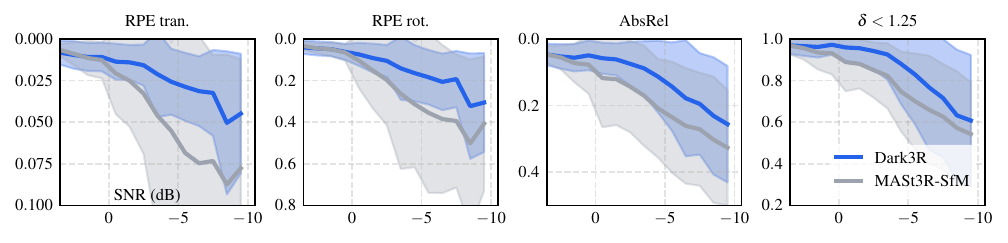}
\vspace{-2.3em}
\caption{Pose prediction assessment. We compare \methodname (blue) against MASt3R-SfM~\cite{duisterhof2025mast3r} (gray) across four pose and depth metrics: relative pose error in translation (RPE~tran.), relative pose error in rotation (RPE~rot.), absolute relative depth error (AbsRel), and the accuracy threshold $\delta < 1.25$.
Each curve shows mean performance with shaded regions indicating standard deviation across scenes.
As image SNR decreases, \methodname maintains lower pose and depth errors and higher reconstruction accuracy compared to MASt3R-SfM.}
\label{fig:results-pose-plots}
\vspace{-1em}
\end{figure*}

\begin{figure*}
\includegraphics[width=\textwidth]
{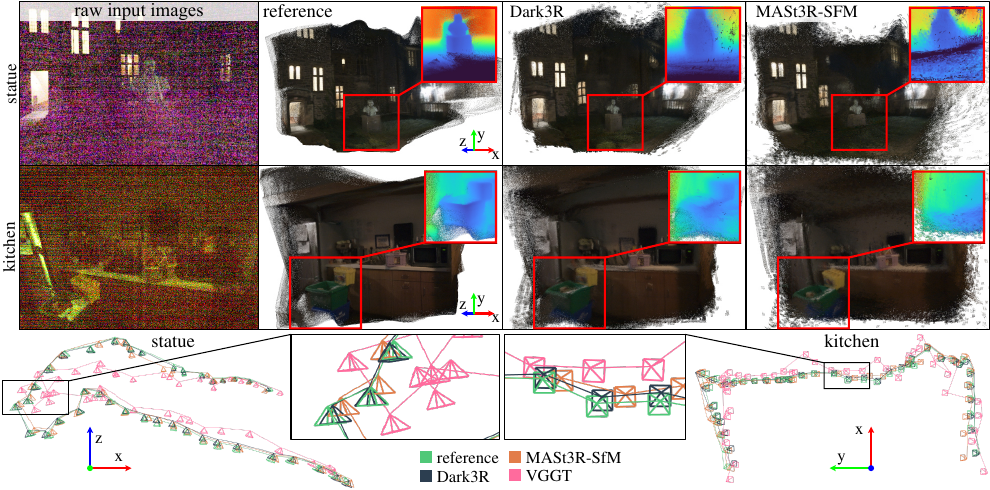}
\vspace{-2em}
\caption{
We compare point clouds and camera poses recovered by \methodname with those estimated by MASt3R-SfM~\cite{duisterhof2025mast3r} on two low-light scenes. \methodname produces more accurate geometry and camera trajectories that better align with a reference solution obtained by running COLMAP~\cite{schonberger2016structure} on well-exposed images captured from the same viewpoints as the noisy inputs.
}
\vspace{-1em}
\label{fig:pose-depth}
\end{figure*}

\section{Dataset}
\label{sec:dataset}
\vspace{-0.3em}
We capture a dataset of multi-view raw images comprising over 100 scenes and 63,000 images using a Sony Alpha~I camera.
The dataset is divided into two subsets: a hand-held-captured dataset without exposure bracketing, and a tripod-captured dataset with bracketing.
Please see Supp.\ Section~\ref*{sec:supp-dataset} for camera capture settings, camera noise model calibration, and a detailed breakdown of scenes and images.

\vspace{-1em}
\paragraph{Handheld-captured dataset.} 
We capture 21,688 well-exposed raw images across 92 distinct indoor scenes with roughly 100--400 images per scene.
We operate the camera in burst mode, which records 10 frames per second.
To use this dataset for training \methodname, we calibrate a Poisson-Gaussian model to the camera and then sample noisy images spanning SNRs from {$-1$ to $-7$\,dB}.

\vspace{-1em}
\paragraph{Tripod-captured (exposure-bracketed) dataset.} 
This dataset contains 41,967 raw images captured across 12 scenes using a tripod-mounted camera.
For each scene, we acquire approximately 400 viewpoints and nine exposures per viewpoint in a bracketed sequence, spaced by 0.7\,EV per step, yielding a dynamic range that spans from well-exposed to extremely low-SNR conditions.  
The lowest two exposures reach mean SNRs below 0\,dB (as low as $-5$\,dB on average and $-10$\,dB at the pixel level), and the measured scene illuminance ranges from 1--15 lux.   
We apply COLMAP~\cite{schonberger2016structure} to the longest exposure in each bracketed sequence and use an image resolution 2048$\times$1408 to provide a reference for pose estimation.

\vspace{-1em}
\paragraph{Training splits.} 
We train on the handheld-captured dataset as described above, and we use seven of the 12 tripod-captured scenes for training.
We hold out five scenes for evaluation. 
For NeRF reconstruction, we train on 90\% of the viewpoints at one exposure level. For NeRF evaluation, we uniformly subsample 10\% of the views held-out viewpoints and translate them by 25x the average inter-camera distance along the camera's forward direction. To derive the corresponding ground truth images, we train an ``oracle'' NeRF on the \textit{clean} training images and render it from the translated held-out viewpoints. 
%The longest exposure in each bracketed sequence is used to provide reference poses using COLMAP.  
For both pose estimation and radiance field reconstruction, training images are downsampled and center-cropped to a resolution of 512$\times$352 pixels from the original resolution of 8640$\times$5760 (50 MP). This 16$\times$ downsampling also improves SNR by a factor of 16 and we report all SNRs on these downsampled images.
\vspace{-0.9em}
\section{Results}
\label{sec:results}
\vspace{-0.6em}
\paragraph{Metrics.}
We evaluate our approach along three dimensions: pose accuracy, depth consistency, and view synthesis photometric quality.  
For pose estimation, we report the absolute translation error (ATE) and the relative pose error broken out into its translational and rotational components  (RPE tran.\ and RPE rot.)~\cite{zhang2025monst3r,zhao2022particlesfm,chen2024leap}, where we treat COLMAP poses estimated from clean images from our exposure-bracketed dataset as the reference.
We evaluate depth consistency using the absolute relative error (AbsRel)~\cite{eigen2014depth} and $\delta < 1.25$, which assesses the percentage of predicted depths within a 1.25 scale factor of the true depth.
% We use per-frame median scaling between the input and reference depth maps~\cite{wang2024dust3r}. % we do not do this right now for depth. We have the scale from the alignment 
% We re-scale the input depth map to match the reference  after aligning the predicted and ground truth camera poses.  
Finally, we use peak signal-to-noise ratio (PSNR), learned perceptual image patch similarity (LPIPS)~\cite{zhang2018unreasonable}, and the structural similarity index measure (SSIM)~\cite{wang2004image} to assess photometric quality.  

\begin{figure*}[ht]
\includegraphics[width=\textwidth]{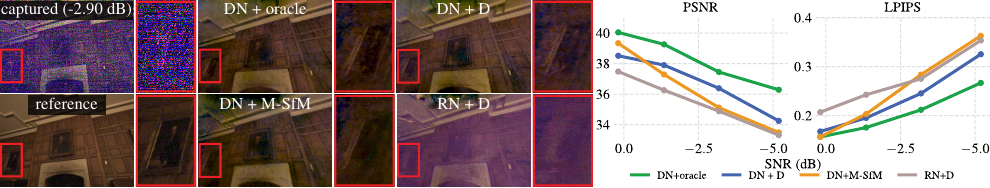}
\vspace{-1em}
\caption{
    We compare novel view synthesis results using two pose estimation methods---\methodname (D) and MASt3R-SfM (M-SfM)~\cite{duisterhof2025mast3r}---and two radiance field reconstruction methods---RawNeRF~\cite{mildenhall2022rawnerf}(RN) and our proposed Dark3R-NeRF (DN). An oracle configuration consisting of Dark3R-NeRF using COLMAP~\cite{schonberger2016structure} poses from well-exposed images serves as an upper bound. 
Our method produces more consistent novel views (insets) and maintains higher quality under increasing noise, as shown in the plots of peak signal-to-noise ratio (PSNR) and LPIPS versus SNR. Please see the project webpage for video results.}
 
\label{fig:results-nerf}
\vspace{-1.5em}
\end{figure*}

\vspace{-1.5em}

% In Fig.~\ref{fig:reb_fig_6}, for each target SNR level ($0$\,dB, $-1.5$\,dB, $-3$\,dB, $-5$\,dB), we select, for each of the five test scenes, the exposure whose average SNR is closest to that level, and then average both SNR and PSNR/LPIPS across scenes. 
% This ensures that each point is computed from the same number of samples.

\paragraph{Baselines.}
We compare against conventional and learned approaches representative of current state-of-the-art structure-from-motion and neural rendering pipelines.  
For pose estimation, we include COLMAP~\cite{schonberger2016structure}, VGGT~\cite{wang2025vggt}, MASt3R~\cite{leroy2024grounding}, and MASt3R-SfM~\cite{duisterhof2025mast3r}.  
% We evaluate all methods on both sRGB and raw image representations.  
All baselines use raw images as input, except for COLMAP and VGGT, which we found to work better after processing the raw images into sRGB.
For radiance field reconstruction, we compare to LE3D \cite{jin2024le3d}, a 3DGS-based \cite{kerbl2023gsplat}
method specifically adapted for noisy raw images, and RawNeRF~\cite{mildenhall2022rawnerf}, which we re-implement in the Nerfacto framework~\cite{nerfstudio} to be consistent with our method.
Please see Supp.\ Section~\ref*{sec:supp-baselines} for implementation details.

\subsection{Pose Estimation}
Fig.~\ref{fig:results-pose-plots} summarizes pose estimation performance and photometric quality versus SNR level across six exposure-bracketed captures of the five held-out scenes in our tripod-captured dataset.
We find that the baselines performance degrades, especially as SNR levels drop below 0 dB. 
Although performance of \methodname also degrades with lower SNR, it does so at a slower rate.

We show additional quantitative results in Table~\ref{tab:pose_comp}, reporting average metric values across the four held-out scenes for a single exposure-bracket setting.
For these scenes, the average image SNR ranges from -4.76 dB to -2.99 dB, and the performance trends are consistent with Fig.~\ref{fig:results-pose-plots}.
The first rows of Table~\ref{tab:pose_comp} are shown for 120 input images per scene, as scaling beyond this requires large GPUs with $>$48 GB VRAM. 
We find that VGGT~\cite{wang2025vggt} and MASt3R-SfM are the most competitive methods to \methodname, with MASt3R-SfM outperforming VGGT. 
Both MASt3R-SfM and \methodname can scale up to 500 input images (``full'') without a large GPU (required by VGGT). 
We show results in this setting (Table~\ref{tab:pose_comp}, bottom rows); we find that average pose accuracy worsens slightly as we increasingly rely on bundle adjustment to reconcile the estimated poses rather than the network's learned prior.

Qualitative comparisons for the same low-SNR regime are shown in Fig.~\ref{fig:pose-depth}.  
Reconstructed scenes using poses estimated by \methodname exhibit better geometric details, and the poses more closely match the reference. 

% Preamble (no new packages needed)
\newcommand{\centercell}[1]{\parbox[c]{\linewidth}{\centering #1}}
\newcommand{\leftcell}[1]{\parbox[c]{\linewidth}{\raggedright #1}} % if you want Method left-aligned

% chapel: -3.51
% statue: -2.99
% fireplace: -4.76
% kitchen: -4.20
% mean: -3.87

\begin{table}
\captionof{table}{\textbf{Camera pose estimation results.} using 120 sequential images (top rows) or the full set (330 to 500 images) from each scene (bottom rows). The test scenes have an average SNR of -3.87 dB. We report pose estimation errors after Sim(3) alignment with COLMAP poses computed on the longest exposure in the exposure bracket. We also provide 3D reconstruction metrics by running depth-supervised Nerfacto~\cite{nerfstudio} with estimated camera poses paired with clean sRGB images to offer better intuition on how the pose errors relate to reconstruction quality. We test each pose estimation method on both sRGB and raw inputs and report the best result. } 
    \label{tab:pose_comp}
    \vspace{-0.5em}
    \centering
    \setlength{\tabcolsep}{2pt}
    \resizebox{\columnwidth}{!}{% optional; remove if you want exact widths
    % 7 columns: Method | Input | ATE | RPE trans | RPE rot | PSNR | AbsRel
\begin{tabular}{l l c c c c c}
\toprule
\multicolumn{2}{c}{} &
\multicolumn{3}{c}{\textbf{Camera pose error}} &
\multicolumn{2}{c}{\textbf{3D reconstruction}} \\
\cmidrule(lr){3-5}\cmidrule(lr){6-7}
\textbf{Method} & \textbf{Input} & \textbf{ATE}\,$\downarrow$ & \textbf{RPE T}\,$\downarrow$ & \textbf{RPE R}\,$\downarrow$ & \textbf{AbsRel}\,$\downarrow$ & \textbf{$\delta < 1.25$}\,$\uparrow$ \\
\midrule
COLMAP~\cite{schonberger2016structure} & sRGB & 0.669 & 0.155 & 1.644 & 0.638 & 54.38 \\
MASt3R~\cite{leroy2024grounding} & raw & 0.787 & 0.472 & 2.802 & 0.318 & 39.66 \\
VGGT~\cite{wang2025vggt} & sRGB & \cellcolor{tabthird}0.252 & \cellcolor{tabthird}0.216 & \cellcolor{tabthird}1.047 & \cellcolor{tabthird}0.232 & \cellcolor{tabthird}63.28 \\
MASt3R\text{-}SfM~\cite{duisterhof2025mast3r} & raw & \cellcolor{tabsecond}0.088 & \cellcolor{tabsecond}0.038 & \cellcolor{tabsecond}0.201 & \cellcolor{tabsecond}0.196 & \cellcolor{tabsecond}79.39 \\
\methodname & raw & \cellcolor{tabfirst}0.050 & \cellcolor{tabfirst}0.020 & \cellcolor{tabfirst}0.121 & \cellcolor{tabfirst}0.091 & \cellcolor{tabfirst}93.14 \\
\midrule
MASt3R\text{-}SfM (Full) & raw & \cellcolor{tabsecond}0.206 & \cellcolor{tabsecond}0.039 & \cellcolor{tabsecond}0.218 & \cellcolor{tabsecond}0.194 & \cellcolor{tabsecond}74.95 \\
\methodname (Full) & raw & \cellcolor{tabfirst}0.139 & \cellcolor{tabfirst}0.019 & \cellcolor{tabfirst}0.121 & \cellcolor{tabfirst}0.093 & \cellcolor{tabfirst}92.33 \\
\bottomrule
\vspace{-3em}
\end{tabular}

}
\end{table}

\vspace{-0.4em}
\subsection{Radiance Field Reconstruction}
Fig.~\ref{fig:results-nerf} compares novel view synthesis performance using two pose estimation methods (Dark3R and MASt3R-SfM) and three neural reconstruction methods (RawNeRF, LE3D, our proposed Dark3R-NeRF). For each target SNR  ($0$\,dB, $-1.5$\,dB, $-3$\,dB, $-5$\,dB), we select, for each of the five test scenes, the exposure setting whose average SNR is closest to the target, and then we average both SNR and PSNR/LPIPS across all scenes. 
The oracle configuration, which combines Dark3R-NeRF with MASt3R-SfM poses and depth from well-exposed reference images, serves as an upper bound on achievable quality.
Given a noisy input raw image sequence, Dark3R-NeRF maintains novel views with more detail than using RawNeRF or LE3D for reconstruction or using MASt3R-SfM for pose estimation.

We observe misalignment between our reconstructed images and the reference images due to temporal noise and sensor-specific per-channel scaling in the low-light regime. We compute a per-channel scale and shift using the median between reconstructed and reference image pairs—inspired by alignment strategies in monocular depth estimation~\cite{ranftl2020midas}—and then evaluate PSNR on the aligned results. We plot the average PSNR versus image SNR across all four scenes and exposure-bracketed captures, which shows our approach achieves consistently lower pose and depth errors and higher reconstruction quality as SNR decreases. 
Additional video comparisons are included in the project webpage.
We see similar trends in the qualitative results shown in Table~\ref{tab:nerf_comp}, where we assess performance on the held-out dataset for the same exposure-bracket setting as Table~\ref{tab:pose_comp}.

\begin{table}[t!]
    \centering
    \vspace{-0.75em}
    \caption{\textbf{View synthesis results.} 
    We assess photometric quality resulting from use of different methods for view synthesis on low-SNR images and camera pose estimation.
    We also compare to an ``oracle'' that takes camera poses estimated from clean image data. Our approach for view synthesis \methodname-NeRF outperforms other techniques when paired with \methodname for pose estimation. 
    % The reference images and depth are rendered after applying Nerfacto~\cite{nerfstudio} to the clean raw images and corresponding COLMAP poses.
    }
    \vspace{-0.3em}
    \label{tab:nerf_comp}
    \setlength{\tabcolsep}{2pt}
    % \resizebox{\columnwidth}{!}{%

\footnotesize

\begin{tabular}{l c c c c }
\toprule
\textbf{Method} &
\textbf{Camera poses} &
\textbf{PSNR}\,$\uparrow$ & \textbf{SSIM}\,$\uparrow$ & \textbf{LPIPS}\,$\downarrow$ \\
\midrule
\methodname -NeRF & MASt3R-SfM\cite{duisterhof2025mast3r} & 34.60 & 0.835 & 0.308 \\
RawNeRF \cite{mildenhall2022rawnerf} & \methodname & 34.24 & 0.848 & \cellcolor{tabthird}0.291 \\
LE3D \cite{jin2024le3d} & \methodname & \cellcolor{tabthird}35.77 & \cellcolor{tabsecond}0.878 & 0.339 \\
% \midrule
\methodname -NeRF & \methodname & \cellcolor{tabsecond}36.17 & \cellcolor{tabthird}0.866 & \cellcolor{tabsecond}0.257 \\\midrule
\methodname -NeRF & Oracle & \cellcolor{tabfirst}37.16 & \cellcolor{tabfirst}0.882 & \cellcolor{tabfirst}0.228 \\
\bottomrule
\vspace{-2em}
\end{tabular}
% these are at a lower light level - do not use 
% \methodname -NeRF & \methodname & 36.88 & 0.868 & 0.299 \\

% \methodname -NeRF & MASt3R-SfM & 36.37 & 0.860 & 0.307 \\
% \bottomrule
\end{table}

% \midrule
% Raw 3DGS & MASt3R-SfM & X & X & X \\
% RawNeRF \cite{mildenhall2022rawnerf} & MASt3R-SfM & X & X & X \\

% Raw 3DGS & Ground truth & X & X & X \\
% RawNeRF \cite{mildenhall2022rawnerf} & Ground truth & 34.68 & 0.884 & 0.321 \\

% \midrule

%%% OLD TABLE: 
    % \begin{tabular}{l c c c c c}
    %     \toprule
    %     & \multicolumn{3}{c}{\textbf{Photometric }} & \multicolumn{2}{c}{\textbf{Depth }} \\
    %     \cmidrule(lr){2-4}\cmidrule(lr){5-6}
    %     \textbf{Method} &
    %     \textbf{PSNR}\,$\uparrow$ &
    %     \textbf{SSIM}\,$\uparrow$ &
    %     \textbf{LPIPS}\,$\downarrow$ &
    %     \textbf{AbsRel}\,$\uparrow$ &
    %     \textbf{Metric 2}\,$\uparrow$ \\
    %     \midrule
    %     Raw 3DGS &
    %     X & X & X & X & X \\
    %     RawNeRF \cite{mildenhall2022rawnerf} &
    %     X & X & X & X & X \\
    %     \midrule
    %     \methodname &
    %     X &
    %     X &
    %     X &
    %     X &
    %     X \\
    %     \bottomrule
    % \end{tabular}}

% \vspace{-0.2em}
\subsection{Ablation Study}
Table~\ref{tab:pose_ablate} reports pose estimation errors under different fine-tuning configurations.  
\textbf{(1)} Using LoRA~\cite{hu2022lora} vs.\ full fine-tuning consistently improves both absolute and relative pose accuracy.  
\textbf{(2)} Operating directly on raw images (instead of sRGB) reduces ATE and RPE errors by preserving linear sensor responses, and \textbf{(3)} supervising with clean sRGB pairs performs slightly worse than the using clean raw image pairs (Eq.~\ref{eq:loss}).  
\textbf{(4--5)} Supervision with only ``simulated data'' works reasonably well (i.e., using our handheld-captured dataset and adding simulated noise); however, performance is improved further by using clean--raw capture pairs (``captured data'') or using both (``proposed''). 
\textbf{(6--8)} We also assess the impact of different fine-tuning schemes, such as only fine-tuning the encoder $\noisyenc$ (and removing $\corrmap$ and $\noisycorrmap$ from the loss function); fine-tuning $\noisyenc$ and $\noisydec$ with the same loss function as Eq.~\ref{eq:loss}; and removing $\mathcal{L}_\text{clean}$ from the loss.
We find that these different fine-tuning schemes only modestly impact the performance.

Table~\ref{tab:recon_ablate} presents an ablation of \methodname{}’s radiance field reconstruction components.  
We find that each component: depth supervision, black level clipping, and stochastic preconditioning are all important to performance. 

Finally, Supp.\ Section~\ref{sec:supp-results} provides additional results evaluating \methodname on clean images (Supp.\ Table~\ref{tab:reb_pose_clean}) and with or without using camera intrinsics (Supp.\ Table \ref{tab:reb_pose_unknown_intrinsics}). 
% We find that each component: depth supervision, black level clipping, and stochastic preconditioning are all important to the performance of the proposed method. 

\subsection{Applications.}
\paragraph{Generalization to other cameras.}
We evaluate \methodname\ on an additional dataset captured by an iPhone 16 without retraining the model.
Quantitative results in Supp.\ Table~\ref{tab:supp-iphone} show that \methodname\ generalizes well to the different noise characteristics of this sensor.  

\vspace{-0.7em}
\paragraph{High dynamic range rendering.}
Finally, we demonstrate that \methodname supports high dynamic range (HDR) rendering directly from raw inputs (see project webpage).  
By operating in the linear sensor irradiance domain, our approach preserves the full dynamic range of the sensor response and enables tone-mapped reconstructions comparable to those produced by RawNeRF~\cite{mildenhall2021nerf}, but in lower SNR conditions where we enable pose estimation.

\vspace{-1.2em}

\begin{table}

\captionof{table}{\textbf{Camera pose estimation ablation study}. 
We ablate a number of design choices, including use of (1) LoRA vs.\ full fine-tuning, (2) using sRGB instead of raw images as input, or (3) supervising only with the former, (4--5) using only simulated or captured data, and (6--8) using different loss functions.
}

\label{tab:pose_ablate}
\centering
\small
\setlength{\tabcolsep}{4pt} % From your code
\renewcommand{\arraystretch}{1.1} % From your code
% We change the column spec from 'cccc ccc' to 'l ccc'
% l = left-aligned text column for description
% ccc = three centered columns for metrics
\footnotesize

{
\begin{tabular}{l c c c}
\toprule
\textbf{Condition} & \textbf{ATE}\,$\downarrow$ & \textbf{RPE T}\,$\downarrow$ & \textbf{RPE R}\,$\downarrow$ \\
\midrule
1. w/o LoRA & 0.476 & 0.074 & 0.512 \\
2. fine-tune \& test w/ sRGB & 0.087 & 0.024 & 0.140 \\
3. supervise w/ clean sRGB & 0.064 & \cellcolor{tabfirst}0.020 & \cellcolor{tabsecond}0.122 \\
4. only simulated data & 0.078 & 0.032 & 0.181 \\
5. only captured data & 0.063 & \cellcolor{tabfirst}0.020 & 0.128 \\
6. fine-tune $\noisyenc$ only & \cellcolor{tabfirst}0.030 & 0.024 & 0.149 \\
7. fine-tune $\noisyenc$ \& $\noisydec$ & \cellcolor{tabsecond}0.049 & \cellcolor{tabfirst}0.020 & 0.124 \\
8. w/o clean loss & 0.051 & \cellcolor{tabfirst}0.020 & \cellcolor{tabfirst}0.121 \\
\midrule
9. \textbf{proposed} & \cellcolor{tabthird}0.050 & \cellcolor{tabfirst}0.020 & \cellcolor{tabfirst}0.121 \\
\bottomrule
\end{tabular}

% smaller ablation table: 
% \begin{tabular}{l c c c}
% \toprule
% \textbf{Condition} & \textbf{ATE}\,$\downarrow$ & \textbf{RPE T}\,$\downarrow$ & \textbf{RPE R}\,$\downarrow$ \\
% \midrule
% 1. no finetuning (MASt3r-SfM) & 0.088 & 0.038 & 0.201  \\
% 2. only simulated data (8100 frames) & \cellcolor{tabthird}0.078 & \cellcolor{tabthird}0.032 & \cellcolor{tabthird}0.181 \\
% 3. only captured data (2700 frames) & \cellcolor{tabsecond}0.063 & \cellcolor{tabfirst}0.020 & \cellcolor{tabsecond}0.128 \\
% \midrule
% 4. \textbf{proposed} & \cellcolor{tabfirst}0.050 & \cellcolor{tabfirst}0.020 & \cellcolor{tabfirst}0.121 \\
% \bottomrule
% \end{tabular}

}

%\vspace{-1em}
\end{table}

\begin{table}
\vspace{-0.6em}
\captionof{table}{ \textbf{View synthesis ablation study.} We assess the impact of depth supervision, stochastic preconditioning, and un-normalized raw inputs on view synthesis. The proposed approach (Dark3R-NeRF) performs the best in terms of photometric quality. } 
    \label{tab:recon_ablate}
    \vspace{-0.5em}
\centering
\footnotesize                                   % modest font shrink (legible in print)
\setlength{\tabcolsep}{4pt}              % tighter columns
\renewcommand{\arraystretch}{1.1}        % a little more row height
\resizebox{\columnwidth}{!}{% ensures one-column width

\begin{tabular}{l ccc}
\toprule
\textbf{Condition} &
\textbf{PSNR}\,$\uparrow$ & \textbf{SSIM}\,$\uparrow$ & \textbf{LPIPS}\,$\downarrow$ \\
\midrule
1. w/o depth supervision & 34.91 & \cellcolor{tabthird}0.858 & 0.307 \\
2. w/ black level clipping & \cellcolor{tabthird}34.98 & 0.856 & \cellcolor{tabthird}0.267 \\
3. w/o stochastic preconditioning & \cellcolor{tabsecond}36.05 & \cellcolor{tabsecond}0.857 & \cellcolor{tabsecond}0.262 \\
\midrule
\textbf{proposed} & \cellcolor{tabfirst}36.17 & \cellcolor{tabfirst}0.866 & \cellcolor{tabfirst}0.257 \\
\bottomrule
\vspace{-3em}
\end{tabular}

}\textbf{ }
\end{table}

\section{Concluding Remarks}
\methodname opens up new possibilities for SfM in low-light regimes where prior methods fail. 
Our results suggest multiple promising avenues for future research. 
One direction is to extend our framework toward feed-forward prediction using large-scale architectures such as VGGT, though this likely requires adapting their auxiliary encoders, such as DINO~\cite{oquab2023dinov2}, to be compatible with raw, low-SNR images. 
Another direction is to build on recent work on dynamic 3D reconstruction~\cite{li2025megasam,jin2025stereo4d,wang2025continuous,zhang2025monst3r}, potentially enabling SfM on dynamic scenes captured in the dark. 
Beyond these 
extensions, integrating generative priors~\cite{szymanowicz2025bolt3d,bahmani2025lyra,gao2024cat3d} could further improve robustness to extreme darkness. 
Overall, we believe this line of research lays the foundation for robust, data-driven 3D understanding under conditions traditionally considered inaccessible to passive vision.

\section*{Acknowledgments}
DBL and KNK acknowledge support from Sony Corporation and from NSERC under the RGPIN and Alliance programs. DBL also acknowledges support from the Canada Foundation for Innovation and the Ontario Research Fund. 

{
    \small
    \bibliographystyle{ieeenat_fullname}
    \bibliography{main}
}

\clearpage
\setcounter{page}{1}
\setcounter{section}{0}
\renewcommand{\thesection}{S\arabic{section}}
\setcounter{figure}{0}
\renewcommand{\thefigure}{S\arabic{figure}}
\setcounter{table}{0}
\renewcommand{\thetable}{S\arabic{table}}

\maketitlesupplementary

We provide supplemental implementation details and results here, and \textbf{\textit{video results and comparisons to baselines are provided in the project webpage.}}
% --- Full Width 4-Column Table ---
\begin{table*}[ht!]
  \centering
  \caption{\textbf{Dataset Statistics.} Summary of the 92 scenes from our handheld-captured dataset used for simulated low-light finetuning. We break down the number of images along with a description per scene. }
  \label{tab:scene_stats_full}
  \scriptsize
  \setlength{\tabcolsep}{5pt} 
  \begin{tabular*}{\textwidth}{@{\extracolsep{\fill}} lr | lr | lr | lr }
    \toprule
    \textbf{Scene} & \textbf{Num. Images} & \textbf{Scene} & \textbf{Num. Images} & \textbf{Scene} & \textbf{Num. Images} & \textbf{Scene} & \textbf{Num. Images} \\
    \midrule
    Bookstore 1 & 249 & Furniture Store 9 & 248 & Kitchen 1 & 172 & Old College 5 & 492 \\
    Bookstore 2 & 196 & Furniture Store 10 & 336 & Lab 1 & 191 & Old College 6 & 212 \\
    Bookstore 3 & 231 & Furniture Store 11 & 225 & Lecture Hall 1 & 270 & Old College 7 & 251 \\
    Catering 1 & 184 & Furniture Store 12 & 173 & Library 1 & 409 & Parking Lot 1 & 216 \\
    Classroom 1 & 242 & Furniture Store 13 & 154 & Library 2 & 230 & Parking Lot 2 & 245 \\
    Classroom 2 & 302 & Furniture Store 14 & 284 & Library 3 & 235 & Parking Lot 3 & 309 \\
    Conference Room 1 & 156 & Furniture Store 15 & 55 & Library 4 & 159 & Parking Lot 4 & 192 \\
    Department Store 1 & 207 & Furniture Store 16 & 278 & Library 5 & 151 & Parking Lot 5 & 511 \\
    Department Store 2 & 166 & Furniture Store 17 & 232 & Library 6 & 201 & Parking Lot 6 & 208 \\
    Department Store 3 & 185 & Furniture Store 18 & 188 & Library 7 & 292 & Parking Lot 7 & 257 \\
    Department Store 4 & 162 & Furniture Store 19 & 223 & Library 8 & 191 & Parking Lot 8 & 227 \\
    Department Store 5 & 307 & Furniture Store 20 & 186 & Library 9 & 258 & Storage Room 1 & 270 \\
    Department Store 6 & 222 & Furniture Store 21 & 196 & Lobby 1 & 175 & Storage Room 2 & 178 \\
    Department Store 7 & 201 & Furniture Store 22 & 205 & Mini Golf 1 & 212 & Storage Room 3 & 175 \\
    Department Store 8 & 153 & Furniture Store 23 & 242 & Museum 1 & 192 & Subway Station 1 & 318 \\
    Furniture Store 1 & 236 & Furniture Store 24 & 137 & Museum 2 & 166 & Subway Station 2 & 105 \\
    Furniture Store 2 & 205 & Grocery Store 1 & 340 & Office 1 & 195 & Utilities 1 & 242 \\
    Furniture Store 3 & 208 & Grocery Store 2 & 195 & Office 2 & 242 & Workshop 1 & 271 \\
    Furniture Store 4 & 337 & Grocery Store 3 & 417 & Office 3 & 311 & Workshop 2 & 295 \\
    Furniture Store 5 & 243 & Grocery Store 4 & 206 & Old College 1 & 191 & Workshop 3 & 333 \\
    Furniture Store 6 & 263 & Grocery Store 5 & 260 & Old College 2 & 222 & Workshop 4 & 337 \\
    Furniture Store 7 & 241 & Grocery Store 6 & 155 & Old College 3 & 229 & Workshop 5 & 152 \\
    Furniture Store 8 & 206 & Hardware Space 1 & 178 & Old College 4 & 400 & Workshop 6 & 383 \\
    \bottomrule
  \end{tabular*}
\end{table*}

\begin{figure*}[ht]
\centering
\includegraphics[width=0.95\textwidth]{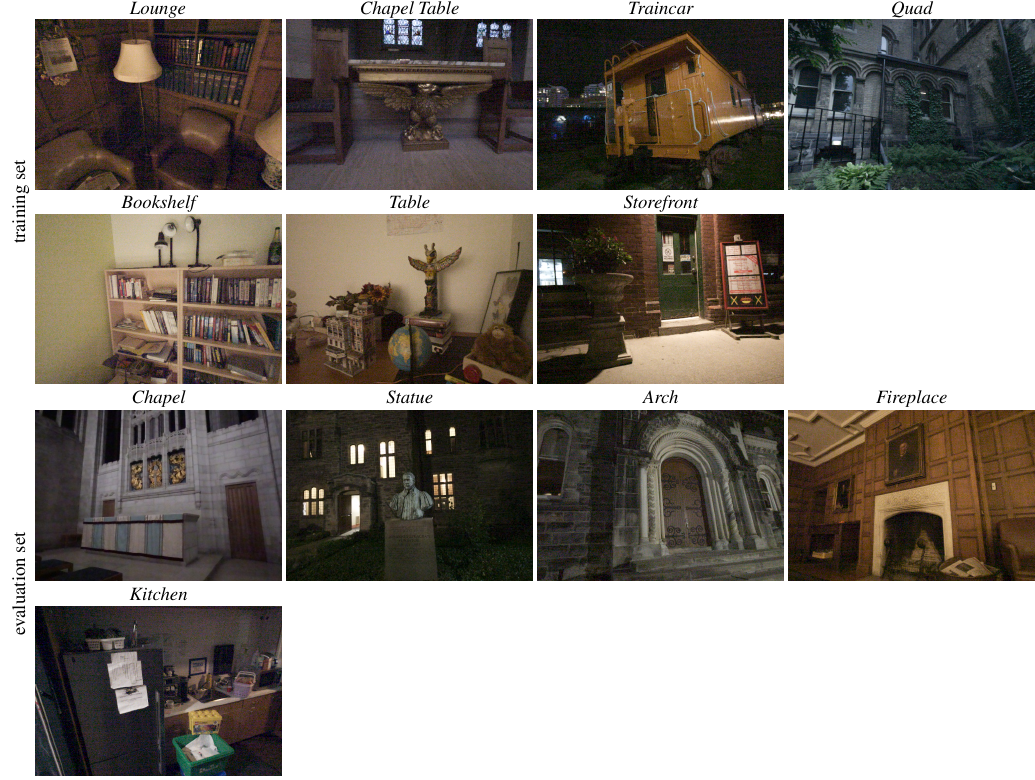}
\caption{Scenes included in the tripod-captured dataset. We capture 12 different scenes and use seven for training and five for evaluation.}
\label{fig:supp-dataset-overview}
\end{figure*}

\begin{figure*}[ht]
\centering
\includegraphics[width=0.95\textwidth]{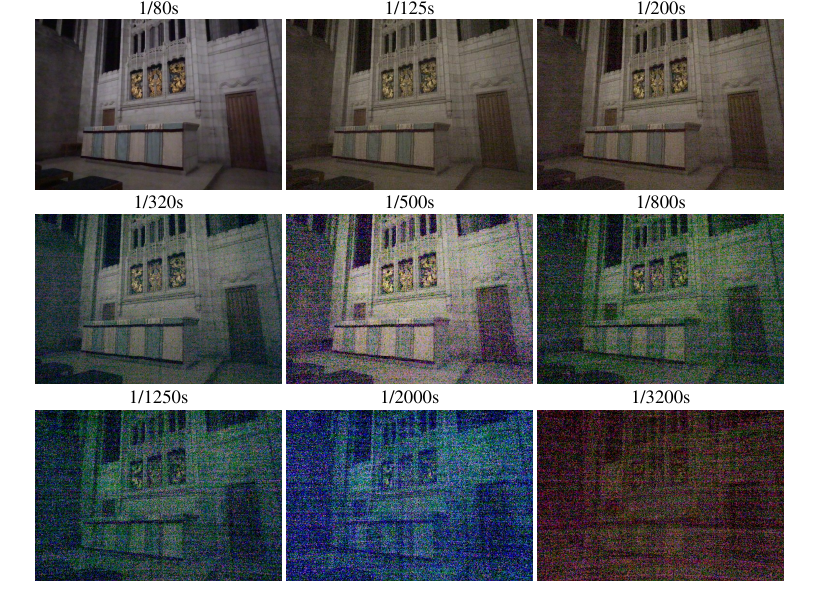}
\caption{
We visualize the different exposures included per scene in our dataset with shutter speeds indicated above each image. We capture nine different exposures for each scene across a wide range of SNRs. Images are contrast-stretched to facilitate visualization.}
\label{fig:supp-dataset-overview}
\end{figure*}

\begin{figure*}[ht]
\vspace{-0.5em}
\includegraphics[width=\textwidth]{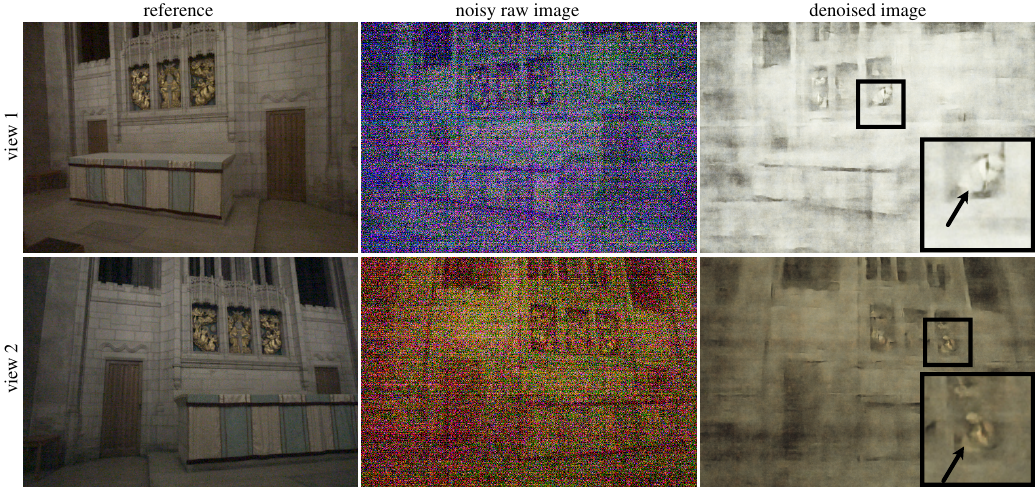}
\caption{We illustrate the result of applying the 2D denoiser LED~\cite{jin2023led} to captured images in our dataset. Inspecting the output of the denoiser shows that the appearance of image features is not multi-view consistent (see insets, arrows), which complicates feature matching used by conventional structure-from-motion pipelines.}
\label{fig:supp-denoising-feature-matching}
\vspace{-1em}
\end{figure*}

\begin{figure*}[ht]
\includegraphics[width=\textwidth]{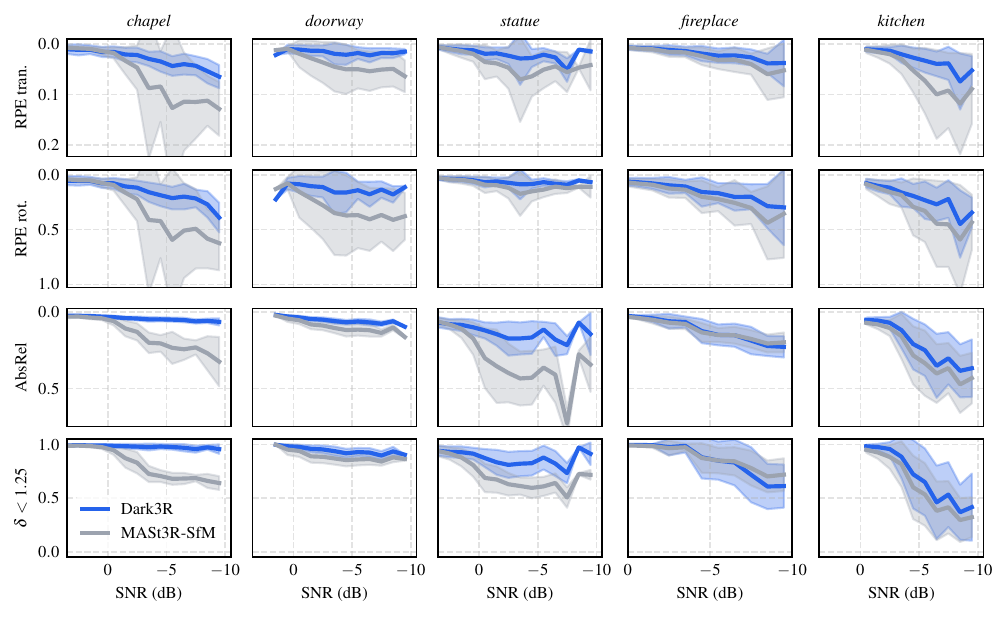}
\caption{Pose prediction assessment for all evaluated scenes. We compare Dark3R (blue) against MASt3R-SfM~\cite{duisterhof2025mast3r}  (gray) across four pose and depth metrics:
relative pose error in translation (RPE tran.), relative pose error in rotation (RPE rot.), absolute relative depth error (AbsRel), and the accuracy threshold $\delta$$<$1.25. Each curve shows mean performance with shaded regions indicating standard deviation across scenes. 
As image SNR decreases, Dark3R generally maintains lower pose and depth errors and higher reconstruction accuracy compared to MASt3R-SfM.}
\label{fig:supp-pose-vs-snr}
\end{figure*}

\begin{figure*}[h]
\centering
\includegraphics[width=.98\textwidth]{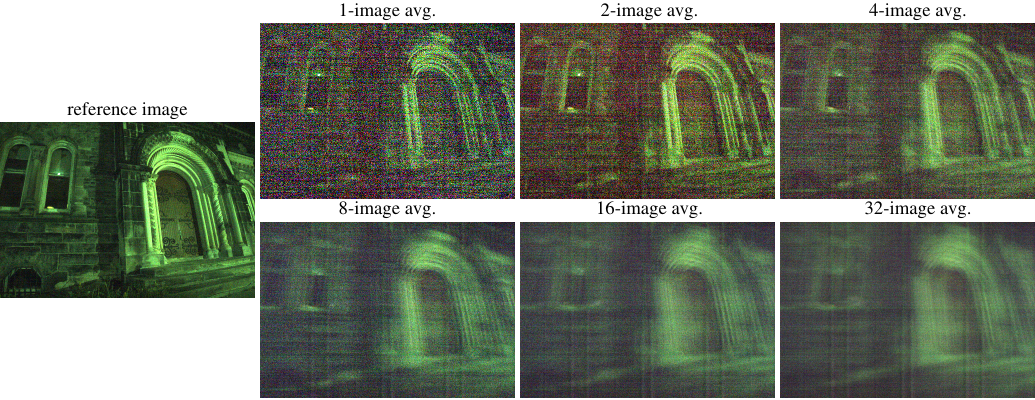}
\caption{Visualization of naive averaging on a captured scene. We average an increasing number of sequentially captured images together to perform denoising, but this comes at the cost of blurring the image. Hence, the dataset contains a non-trivial amount of disparity between the captured images, and the amount of noise makes it challenging to apply conventional burst denoising techniques that require image alignment.}
\label{fig:supp-results-avg}
\end{figure*}

\begin{figure*}[ht]
\includegraphics[width=\textwidth]{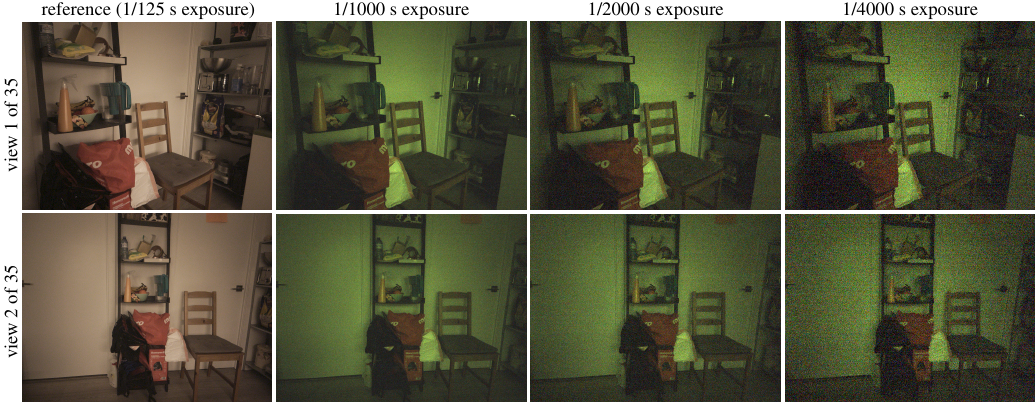}
\caption{iPhone dataset. We test the generalization capabilities of our model by testing it on a multi-view raw image dataset captured by an iPhone 16. We capture this scene at four different exposure settings (including a reference, well-exposed capture), resulting in different signal-to-noise ratios. As shown in Table~\ref{tab:supp-iphone}, the \methodname outperforms the baseline even when applied to this dataset.}
\label{fig:supp-iphone}
\end{figure*}

\section{Supplemental Implementation Details}
\label{sec:supp-implementation-details}

% \begin{figure*}[ht]
% \includegraphics[width=\textwidth]{figures/supp-dataset}
% \caption{Table describing the dataset.}
% \label{fig:supp-dataset}
% \end{figure*}

\subsection{Datasets}
\label{sec:supp-dataset}

\paragraph{Handheld-captured dataset.}
We fine-tune the model on a dataset comprising 21,688 well-exposed raw images captured across 92 distinct scenes, with approximately 100--400 images per scene (or about 200 on average---see Table~\ref{tab:scene_stats_full}).  
All images are captured using a Sony Alpha~I camera in handheld burst mode at 10~frames per second while moving around each scene to ensure significant viewpoint overlap between adjacent frames.  
The exposure time is fixed to $1/200$\,s to avoid motion blur, while the ISO is adjusted to achieve an exposure value (EV) of approximately $-0.3$ relative to the metered exposure, ensuring that no regions are over-exposed.  
The aperture is set to an $f$-number of at least $f/11$ to maximize depth of field, and decreased when necessary to maintain the target exposure time. After downsampling, our captures are clean and exhibit minimal noise. 
% The ISO is kept below~6,400 for all captures.  

We use a calibrated Poisson–Gaussian noise model for the camera to synthesize noisy raw images from the clean captures.
For each clean image, we sample a random scaling factor to determine the mean pixel intensity, then compute the corresponding noise variance according to the noise model.
Poisson-Gaussian noise is then added to produce images spanning a target range of signal-to-noise ratios (SNRs) between $-1$ and $-7$ dB.
This noise synthesis is performed on the fly during fine-tuning, enabling continuous variation in brightness and noise conditions.
All images, together with full capture metadata---including lens model, ISO, aperture, exposure time, and noise parameters---will be publicly released.

The noise model is calibrated following the procedure of Plötz and Roth~\cite{plotz2017benchmarking}.
For each RGB channel of the Sony Alpha 1’s color filter array, we capture GretagMacbeth ColorChecker images at three exposure values and compute the mean–variance pairs over all homogeneous patches. Fitting a linear model to these measurements yields the noise parameters for each channel.

% We analyze the influence of dataset size on model performance in Sec.~\ref{sec:results}.

% --- The Table ---
% [ht] is a placement specifier: "here" or "top" of the page.
\begin{table}[t]
  \centering
  \caption{\textbf{Dataset scenes.} A summary of the scenes used in our dataset, detailing the number of images, environment type, and train/test splits. Each scene contains nine different exposures with ground truth camera poses. }
  \label{tab:dataset_scenes}
  
  % Set column separation (padding). Your example used 2pt, 4-6pt is also common.
  \setlength{\tabcolsep}{6pt} 
  
  % Use footnotesize as in your example
  \footnotesize

  % {l c c c} specifies column alignment:
  % l = left-aligned
  % c = center-aligned
  \begin{tabular}{l c c c}
    \toprule
    \textbf{Scene} & \textbf{Num Images} & \textbf{Environment} & \textbf{Split} \\
    \midrule
    % --- Training Scenes ---
    Lounge & 430 & Indoors & Train \\
    Chapel Table & 420 & Indoors & Train \\
    Traincar & 400 & Outdoors & Train \\
    Quad & 400 & Outdoors & Train \\
    Bookshelf & 361 & Indoors & Train \\
    Table & 332 & Indoors & Train \\
    Storefront & 330 & Outdoors & Train \\
    \midrule % Separate Train and Test splits
    % --- Test Scenes ---
    Chapel & 500 & Indoors & Test \\
    Statue & 400 & Outdoors & Test \\
    Arch & 370 & Outdoors & Test \\
    Fireplace & 360 & Indoors & Test \\
    Kitchen & 360 & Indoors & Test \\
    \bottomrule
  \end{tabular}
\end{table}

\paragraph{Tripod-captured (exposure-bracketed) dataset.}
We further evaluate our approach on a captured low-light dataset consisting of 12 scenes (see Fig.~\ref{fig:supp-dataset-overview}), each containing 300--500 viewpoints recorded using the same Sony Alpha~I camera (see Table~\ref{tab:dataset_scenes}).  
For each scene, we acquire nine exposures per viewpoint in a bracketed sequence, spaced by 0.7\,EV per step, yielding a dynamic range that spans from well-exposed to extremely low-SNR conditions.  
The lowest two exposures reach mean SNRs below 0\,dB (as low as $-5$\,dB on average and $-10$\,dB at the pixel level).  
All captures are performed on a tripod using an aperture of $f/22$ for maximum depth of field and an ISO of 102{,}400, with exposure times adjusted according to the measured scene illuminance (typically 1–15\,lux) to maintain consistent brightness across scenes.  
The longest exposure in each bracketed sequence serves as a reference for pose estimation: we compute ground-truth camera poses using COLMAP on these well-exposed images.  
For radiance field reconstruction, we train on 90\% of the frames and hold out 10\% for evaluation as described in the main text.

% Preamble (no new packages needed)

% chapel: -3.51
% statue: -2.99
% fireplace: -4.76
% kitchen: -4.20
% mean: -3.87
\begin{table}
\captionof{table}{\textbf{Camera pose estimation using a 2D denoiser.} We first process raw noisy images using the 2D denoiser LED \cite{jin2023led} before using those images as input for MASt3R-SfM \cite{duisterhof2025mast3r}. While using a 2D denoiser improves performance slightly compared to the baseline of providing raw input images to MASt3R-SfM, it performs much worse than \methodname, which is trained to operate directly on noisy raw images. These results were computed on the chapel scene using 120 images. } 
    \label{tab:pose_denoiser}
    \vspace{-0.5em}
    \centering
    \setlength{\tabcolsep}{2pt}
    \resizebox{\columnwidth}{!}{% optional; remove if you want exact widths
    % 7 columns: Method | Input | ATE | RPE trans | RPE rot | PSNR | AbsRel
\begin{tabular}{l c c c c }
\toprule
\textbf{Method} & \textbf{Input}   & \textbf{ATE}\,$\downarrow$ & \textbf{RPE T}\,$\downarrow$ & \textbf{RPE R}\,$\downarrow$ \\
\midrule

MASt3R\text{-}SfM \cite{duisterhof2025mast3r} &  raw   & \cellcolor{tabthird}0.080 & \cellcolor{tabthird}0.061 & \cellcolor{tabthird}0.289\\

MASt3R\text{-}SfM \cite{duisterhof2025mast3r} & denoised raw   & \cellcolor{tabsecond}0.048 & \cellcolor{tabsecond}0.048 & \cellcolor{tabsecond}0.238 \\

\methodname  & raw & \cellcolor{tabfirst}0.038 & \cellcolor{tabfirst}0.028 & \cellcolor{tabfirst}0.146 \\
\bottomrule
\end{tabular}

}
\end{table}
\begin{table}
\captionof{table}{
\textbf{Camera pose estimation results by scene.} We report results for 120 sequential images (top rows) or the full set (bottom rows) from each scene, grouped by scene name. Metrics include pose estimation errors after Sim(3) alignment and 3D reconstruction quality.
}
\label{tab:pose_comp_unified_final}
\vspace{-0.5em}
\centering
\setlength{\tabcolsep}{2pt}
\resizebox{\columnwidth}{!}{%
\begin{tabular}{l l l c c c c c}
\toprule
\multicolumn{3}{c}{\textbf{Method/Input}} &
\multicolumn{3}{c}{\textbf{Camera pose error}} &
\multicolumn{2}{c}{\textbf{3D reconstruction}} \\
\cmidrule(lr){4-6}\cmidrule(lr){7-8}
\textbf{Scene} & \textbf{Method} & \textbf{Input} & \textbf{ATE}\,$\downarrow$ & \textbf{RPE T}\,$\downarrow$ & \textbf{RPE R}\,$\downarrow$ & \textbf{AbsRel}\,$\downarrow$ & \textbf{$\delta < 1.25$}\,$\uparrow$ \\
\midrule
\multirow{7}{*}{\centering \textbf{Chapel}} & COLMAP~\cite{schonberger2016structure} & sRGB & \cellcolor{tabthird}0.187 & \cellcolor{tabthird}0.250 & \cellcolor{tabthird}1.229 & \cellcolor{tabsecond}0.187 & \cellcolor{tabsecond}84.21 \\
 & MASt3R~\cite{leroy2024grounding} & raw & 0.832 & 0.614 & 2.877 & 0.234 & 60.56 \\
 & VGGT~\cite{wang2025vggt} & sRGB & 0.405 & 0.451 & 2.197 & 0.212 & 59.24 \\
 & MASt3R\text{-}SfM~\cite{duisterhof2025mast3r} & raw & \cellcolor{tabsecond}0.080 & \cellcolor{tabsecond}0.061 & \cellcolor{tabsecond}0.289 & \cellcolor{tabthird}0.210 & \cellcolor{tabthird}70.74 \\
 & \methodname & raw & \cellcolor{tabfirst}0.038 & \cellcolor{tabfirst}0.027 & \cellcolor{tabfirst}0.144 & \cellcolor{tabfirst}0.050 & \cellcolor{tabfirst}98.22 \\
\cline{2-8}
 & MASt3R\text{-}SfM (Full) & raw & \cellcolor{tabfirst}0.175 & \cellcolor{tabsecond}0.054 & \cellcolor{tabsecond}0.255 & \cellcolor{tabsecond}0.226 & \cellcolor{tabsecond}57.73 \\
 & \methodname (Full) & raw & \cellcolor{tabsecond}0.206 & \cellcolor{tabfirst}0.025 & \cellcolor{tabfirst}0.135 & \cellcolor{tabfirst}0.053 & \cellcolor{tabfirst}93.99 \\
\midrule
\multirow{7}{*}{\centering \textbf{Statue}} & COLMAP~\cite{schonberger2016structure} & sRGB & 0.377 & \cellcolor{tabthird}0.182 & \cellcolor{tabthird}0.482 & 1.179 & \cellcolor{tabthird}54.33 \\
 & MASt3R~\cite{leroy2024grounding} & raw & 0.790 & 0.483 & 2.526 & \cellcolor{tabthird}0.361 & 26.09 \\
 & VGGT~\cite{wang2025vggt} & sRGB & \cellcolor{tabthird}0.325 & 0.216 & 0.622 & \cellcolor{tabsecond}0.266 & 51.80 \\
 & MASt3R\text{-}SfM~\cite{duisterhof2025mast3r} & raw & \cellcolor{tabsecond}0.081 & \cellcolor{tabsecond}0.044 & \cellcolor{tabsecond}0.111 & 0.381 & \cellcolor{tabsecond}63.34 \\
 & \methodname & raw & \cellcolor{tabfirst}0.054 & \cellcolor{tabfirst}0.020 & \cellcolor{tabfirst}0.066 & \cellcolor{tabfirst}0.166 & \cellcolor{tabfirst}79.10 \\
\cline{2-8}
 & MASt3R\text{-}SfM (Full) & raw & \cellcolor{tabsecond}0.207 & \cellcolor{tabsecond}0.046 & \cellcolor{tabsecond}0.107 & \cellcolor{tabsecond}0.406 & \cellcolor{tabsecond}50.72 \\
 & \methodname (Full) & raw & \cellcolor{tabfirst}0.144 & \cellcolor{tabfirst}0.021 & \cellcolor{tabfirst}0.070 & \cellcolor{tabfirst}0.179 & \cellcolor{tabfirst}82.17 \\
\midrule
\multirow{7}{*}{\centering \textbf{Arch}} & COLMAP~\cite{schonberger2016structure} & sRGB & \cellcolor{tabthird}0.043 & \cellcolor{tabthird}0.062 & \cellcolor{tabthird}0.453 & \cellcolor{tabsecond}0.087 & \cellcolor{tabsecond}94.71 \\
 & MASt3R~\cite{leroy2024grounding} & raw & 0.239 & 0.268 & 1.619 & 0.257 & 64.65 \\
 & VGGT~\cite{wang2025vggt} & sRGB & 0.080 & 0.096 & 0.488 & 0.101 & 87.39 \\
 & MASt3R\text{-}SfM~\cite{duisterhof2025mast3r} & raw & \cellcolor{tabsecond}0.024 & \cellcolor{tabsecond}0.025 & \cellcolor{tabsecond}0.180 & \cellcolor{tabthird}0.088 & \cellcolor{tabthird}88.35 \\
 & \methodname & raw & \cellcolor{tabfirst}0.013 & \cellcolor{tabfirst}0.013 & \cellcolor{tabfirst}0.093 & \cellcolor{tabfirst}0.044 & \cellcolor{tabfirst}96.50 \\
\cline{2-8}
 & MASt3R\text{-}SfM (Full) & raw & \cellcolor{tabsecond}0.068 & \cellcolor{tabsecond}0.045 & \cellcolor{tabsecond}0.352 & \cellcolor{tabsecond}0.100 & \cellcolor{tabsecond}87.93 \\
 & \methodname (Full) & raw & \cellcolor{tabfirst}0.048 & \cellcolor{tabfirst}0.017 & \cellcolor{tabfirst}0.142 & \cellcolor{tabfirst}0.057 & \cellcolor{tabfirst}94.74 \\
\midrule
\multirow{7}{*}{\centering \textbf{Fireplace}} & COLMAP~\cite{schonberger2016structure} & sRGB & 1.443 & \cellcolor{tabthird}0.090 & 3.125 & 0.418 & 24.45 \\
 & MASt3R~\cite{leroy2024grounding} & raw & 0.430 & 0.268 & 1.830 & 0.245 & 55.98 \\
 & VGGT~\cite{wang2025vggt} & sRGB & \cellcolor{tabthird}0.176 & 0.112 & \cellcolor{tabthird}0.881 & \cellcolor{tabfirst}0.083 & \cellcolor{tabsecond}97.57 \\
 & MASt3R\text{-}SfM~\cite{duisterhof2025mast3r} & raw & \cellcolor{tabsecond}0.143 & \cellcolor{tabsecond}0.023 & \cellcolor{tabsecond}0.187 & \cellcolor{tabthird}0.118 & \cellcolor{tabthird}94.58 \\
 & \methodname & raw & \cellcolor{tabfirst}0.106 & \cellcolor{tabfirst}0.018 & \cellcolor{tabfirst}0.146 & \cellcolor{tabsecond}0.090 & \cellcolor{tabfirst}98.64 \\
\cline{2-8}
 & MASt3R\text{-}SfM (Full) & raw & \cellcolor{tabsecond}0.225 & \cellcolor{tabsecond}0.020 & \cellcolor{tabsecond}0.179 & \cellcolor{tabsecond}0.067 & \cellcolor{tabsecond}96.32 \\
 & \methodname (Full) & raw & \cellcolor{tabfirst}0.138 & \cellcolor{tabfirst}0.014 & \cellcolor{tabfirst}0.127 & \cellcolor{tabfirst}0.051 & \cellcolor{tabfirst}97.91 \\
\midrule
\multirow{7}{*}{\centering \textbf{Kitchen}} & COLMAP~\cite{schonberger2016structure} & sRGB & 1.297 & \cellcolor{tabthird}0.191 & 2.933 & 1.320 & 14.20 \\
 & MASt3R~\cite{leroy2024grounding} & raw & 1.289 & 0.521 & 4.186 & \cellcolor{tabthird}0.421 & 7.35 \\
 & VGGT~\cite{wang2025vggt} & sRGB & \cellcolor{tabthird}0.272 & 0.203 & \cellcolor{tabthird}1.048 & 0.500 & \cellcolor{tabthird}20.39 \\
 & MASt3R\text{-}SfM~\cite{duisterhof2025mast3r} & raw & \cellcolor{tabsecond}0.115 & \cellcolor{tabsecond}0.038 & \cellcolor{tabsecond}0.236 & \cellcolor{tabsecond}0.181 & \cellcolor{tabsecond}79.95 \\
 & \methodname & raw & \cellcolor{tabfirst}0.041 & \cellcolor{tabfirst}0.019 & \cellcolor{tabfirst}0.156 & \cellcolor{tabfirst}0.108 & \cellcolor{tabfirst}93.24 \\
\cline{2-8}
 & MASt3R\text{-}SfM (Full) & raw & \cellcolor{tabsecond}0.355 & \cellcolor{tabsecond}0.030 & \cellcolor{tabsecond}0.194 & \cellcolor{tabsecond}0.169 & \cellcolor{tabsecond}82.06 \\
 & \methodname (Full) & raw & \cellcolor{tabfirst}0.158 & \cellcolor{tabfirst}0.018 & \cellcolor{tabfirst}0.132 & \cellcolor{tabfirst}0.124 & \cellcolor{tabfirst}92.85 \\
\bottomrule
\end{tabular}
}
\end{table}
\begin{table}[ht]
\centering
\caption{\textbf{View synthesis results by scene.} We assess photometric quality on low-SNR images. We compare to an ``oracle'' obtained by performing pose estimation on clean images.}
\label{tab:nerf_comp_scenes}
\setlength{\tabcolsep}{2pt}
\resizebox{\columnwidth}{!}{%
\begin{tabular}{l c c c c c}
\toprule
\textbf{Scene} & \textbf{Method} & \textbf{Camera poses} & \textbf{PSNR}\,$\uparrow$ & \textbf{SSIM}\,$\uparrow$ & \textbf{LPIPS}\,$\downarrow$ \\
\midrule 
\multirow{5}{*}{\centering \textbf{Chapel}} & LE3D \cite{jin2024le3d} & \methodname & \cellcolor{tabfirst}38.01 & \cellcolor{tabthird}0.868 & 0.398 \\
 & RawNeRF \cite{mildenhall2022rawnerf} & \methodname & 35.59 & \cellcolor{tabsecond}0.871 & \cellcolor{tabthird}0.347 \\
 & \methodname -NeRF & MASt3R-SfM~\cite{duisterhof2025mast3r} & 36.48 & 0.826 & 0.356 \\
 & \methodname -NeRF & \methodname & \cellcolor{tabthird}37.44 & 0.840 & \cellcolor{tabsecond}0.310 \\
 & \methodname -NeRF & Oracle & \cellcolor{tabsecond}37.87 & \cellcolor{tabfirst}0.873 & \cellcolor{tabfirst}0.270 \\
\midrule
\multirow{5}{*}{\centering \textbf{Statue}} & LE3D \cite{jin2024le3d} & \methodname & \cellcolor{tabthird}25.20 & \cellcolor{tabthird}0.839 & 0.283 \\
 & RawNeRF \cite{mildenhall2022rawnerf} & \methodname & 24.24 & 0.829 & \cellcolor{tabthird}0.255 \\
 & \methodname -NeRF & MASt3R-SfM~\cite{duisterhof2025mast3r} & 24.47 & 0.814 & 0.277 \\
 & \methodname -NeRF & \methodname & \cellcolor{tabsecond}25.68 & \cellcolor{tabsecond}0.851 & \cellcolor{tabsecond}0.240 \\
 & \methodname -NeRF & Oracle & \cellcolor{tabfirst}27.97 & \cellcolor{tabfirst}0.877 & \cellcolor{tabfirst}0.201 \\
\midrule
\multirow{5}{*}{\centering \textbf{Arch}} & LE3D \cite{jin2024le3d} & \methodname & \cellcolor{tabsecond}38.50 & \cellcolor{tabfirst}0.889 & 0.292 \\
 & RawNeRF \cite{mildenhall2022rawnerf} & \methodname & 37.00 & 0.845 & \cellcolor{tabthird}0.275 \\
 & \methodname -NeRF & MASt3R-SfM~\cite{duisterhof2025mast3r} & 35.90 & 0.795 & 0.344 \\
 & \methodname -NeRF & \methodname & \cellcolor{tabthird}38.13 & \cellcolor{tabthird}0.858 & \cellcolor{tabsecond}0.237 \\
 & \methodname -NeRF & Oracle & \cellcolor{tabfirst}38.55 & \cellcolor{tabsecond}0.862 & \cellcolor{tabfirst}0.213 \\
\midrule
\multirow{5}{*}{\centering \textbf{Fireplace}} & LE3D \cite{jin2024le3d} & \methodname & \cellcolor{tabthird}39.35 & \cellcolor{tabfirst}0.898 & 0.408 \\
 & RawNeRF \cite{mildenhall2022rawnerf} & \methodname & 37.57 & 0.834 & 0.306 \\
 & \methodname -NeRF & MASt3R-SfM~\cite{duisterhof2025mast3r} & 38.65 & 0.869 & \cellcolor{tabthird}0.302 \\
 & \methodname -NeRF & \methodname & \cellcolor{tabfirst}41.09 & \cellcolor{tabthird}0.896 & \cellcolor{tabsecond}0.249 \\
 & \methodname -NeRF & Oracle & \cellcolor{tabsecond}40.24 & \cellcolor{tabsecond}0.898 & \cellcolor{tabfirst}0.246 \\
\midrule
\multirow{5}{*}{\centering \textbf{Kitchen}} & LE3D \cite{jin2024le3d} & \methodname & \cellcolor{tabthird}37.81 & \cellcolor{tabsecond}0.897 & 0.312 \\
 & RawNeRF \cite{mildenhall2022rawnerf} & \methodname & 36.81 & 0.860 & 0.272 \\
 & \methodname -NeRF & MASt3R-SfM~\cite{duisterhof2025mast3r} & 37.52 & 0.872 & \cellcolor{tabthird}0.261 \\
 & \methodname -NeRF & \methodname & \cellcolor{tabsecond}38.54 & \cellcolor{tabthird}0.884 & \cellcolor{tabsecond}0.249 \\
 & \methodname -NeRF & Oracle & \cellcolor{tabfirst}41.17 & \cellcolor{tabfirst}0.900 & \cellcolor{tabfirst}0.210 \\
\bottomrule
\end{tabular}
}
\end{table}

% Additionally, for qualitative analysis, we capture several handheld sequences of 300--500 frames per scene using exposure settings that match the three or four shortest exposures in the bracketed set.  
% These handheld captures do not have reference poses but serve to illustrate method performance under more realistic, handheld capture settings.

% Do not uncomment unless we include handheld reconstruction results 

\subsection{Structure From Motion in the Dark}

\paragraph{Input preprocessing.}
The inputs to the model are 14-bit raw sensor measurements, either simulated or captured using a Sony $\alpha$1 mirrorless camera. 
Each captured raw image is center-cropped from the native camera resolution of $8640\times5760$~pixels to $8196\times5632$, then demosaiced by subsampling pixels from each color channel in the Bayer mosaic and averaging the two green channels. 
The demosaiced image is subsequently downsampled by a factor of~8 using OpenCV’s \texttt{INTER\_AREA} interpolation, resulting in a final input resolution of $512\times352$~pixels---matching the maximum resolution supported by MASt3R. 
Finally, the images are normalized to the range~$[0,1]$ in \texttt{float32} using a fixed scale factor of $(2^{14}-1)^{-1}$ based on the sensor’s bit depth.

\paragraph{Data loading.}
At each iteration, the dataloader provides a batch consisting of four images: two high-SNR (clean) and two low-SNR (noisy). 
A clean image is paired with a nearby-captured frame that observes an overlapping region of the scene to form a clean image pair $(\img{1}_{\text{clean}}, \img{2}_{\text{clean}})$. 
The corresponding noisy pair $(\img{1}_{\text{noisy}}, \img{2}_{\text{noisy}})$ is obtained either using simulation or by capturing  images from the same viewpoints using a short exposure. 
This configuration uses roughly 48~GB of GPU memory.

\subsection{View Synthesis in the Dark}
We build \methodname-NeRF on Nerfacto ~\cite{nerfstudio}. Following RawNeRF \cite{mildenhall2022rawnerf}, we supervise and render in linear raw space. To visualize the NeRF renders, we apply an ISP to recover sRGB images. The ISP performs black level subtraction, clipping, scaling, white balance, and gamma correction. The ground truth used to compute quantitative metrics is the clean raw images captured at the longest exposure for the held out viewpoint. 

% \begin{itemize}
% \item   to visualize the nerf renders, we apply an isp to recover srgb images. The isp does black level subtraction, clipping, scaling, gamma correction, and white balance
% \item  ground truth for the nerf is the raw of the clean images from the held out viewpoint and downsampled to 512x352
% \end{itemize}

\subsection{Baseline Implementation Details}
\label{sec:supp-baselines}
For COLMAP and VGGT, the sRGB images are processed through a standard in-camera image signal processor (ISP) including demosaicing via subsampling, black-level subtraction and clipping, white balancing, and gamma correction.  
The raw images are demosaiced by subsampling.

For radiance field reconstruction, we compare against two low-light extensions of common methods.  
We choose to test LE3D \cite{jin2024le3d} as one of our baseline methods. LE3D is a state-of-the-art 3D Gaussian-Splatting~\cite{kerbl2023gsplat} method designed for noisy raw input images. We run their codebase directly.  
Initialization is performed using the refined point cloud produced by \methodname\ during its bundle-adjustment step.

The second method we test is RawNeRF~\cite{mildenhall2022rawnerf}, which we re-implemented within the Nerfacto framework~\cite{nerfstudio} to ensure a consistent backbone with our method.  
It employs the exponential loss proposed in the original RawNeRF paper, along with black-level subtraction and clipping.  
Our method differs by introducing coarse-to-fine optimization, depth supervision, and operating directly on unnormalized raw data. 
All methods are trained for the same number of iterations (90{,}000) for fair comparison.

% \paragraph{Architectural details.}
% \begin{itemize}
%     \item Provide a text description and figure that shows which layers we fine-tune.
% \end{itemize}

\section{Supplemental Results}
\label{sec:supp-results}

\paragraph{2D denoising baseline.}
We show an example of applying a 2D denoiser (LED~\cite{jin2023led}) to a noisy captured scene in Fig.~\ref{fig:supp-denoising-feature-matching}.
Since the denoiser does not preserve image features in a multi-view consistent fashion, it is still challenging to perform pose estimation on the denoised images. 
Specifically, in Table~\ref{tab:pose_denoiser} we find that while MASt3R-SfM performs slightly better in terms of pose accuracy on the denoised images vs. using raw images, Dark3R yields significantly better performance when given the raw images as input. 

\paragraph{Additional pose results.}
In Figure~\ref{fig:supp-pose-vs-snr} we show additional plots of pose and depth accuracy metrics versus image SNR, broken out by scene, on all images from the captured evaluation dataset.
We find that \methodname outperforms MASt3R-SfM in most cases, especially as the SNR decreases.
We see similar trends in the corresponding results shown in Table~\ref{tab:pose_comp_unified_final} across the test scenes. 
These results are calculated for a single exposure-bracket setting in each scene, and the average image SNR per scene ranges from -4.76 dB to -2.99 dB.

\paragraph{Additional view synthesis results.}
We show the results on view synthesis for each scene in the evaluation dataset in Table~\ref{tab:nerf_comp_scenes}, and we use the same single exposure-bracket setting for each scene as in Table~\ref{tab:pose_comp_unified_final}.
We see the same trends as in the main paper---on average using \methodname for pose estimation and Dark3R-NeRF for view synthesis outperforms the baselines. 
We note that in some cases the performance of an oracle (where poses are obtained from the clean images) performs worse than other approaches; however, we attribute this to the fact that the oracle is still trained on noisy images and so there are still artifacts in the reconstruction that can degrade image quality.

\paragraph{Averaging and motion blur in the captured dataset.}
To illustrate that our dataset has significant disparity between views, we show the result of naive averaging of sequentially captured frames for a particular scene (Fig.~\ref{fig:supp-results-avg}).
We observe that while the noise is reduced by averaging images together, this process also introduces significant blur.

\paragraph{Evaluation on the iPhone dataset.}
% Preamble (no new packages needed)

% chapel: -3.51
% statue: -2.99
% fireplace: -4.76
% kitchen: -4.20
% mean: -3.87
\begin{table}
\captionof{table}{\textbf{Generalization to other cameras.} We assess the ability of \methodname to generalize to other types of cameras using images captured on an iPhone 16. We captured a series of low-light multi-view images on an iPhone 16 and tested \methodname against MASt3R-SfM \cite{duisterhof2025mast3r}. Despite not being finetuned with iPhone images, \methodname still outperforms MASt3R-SfM for camera pose estimation accuracy. } 
    \label{tab:supp-iphone}
    \vspace{-0.5em}
    \centering
    \setlength{\tabcolsep}{2pt}
    \resizebox{\columnwidth}{!}{% optional; remove if you want exact widths
    % 7 columns: Method | Input | ATE | RPE trans | RPE rot | PSNR | AbsRel
\begin{tabular}{l c c c c c }
\toprule
\textbf{Method} & \textbf{Input} & \textbf{Exposure (s)} & \textbf{ATE}\,$\downarrow$ & \textbf{RPE T}\,$\downarrow$ & \textbf{RPE R}\,$\downarrow$ \\
\midrule

MASt3R\text{-}SfM \cite{duisterhof2025mast3r} & raw & 1/1000 & \cellcolor{tabsecond}0.105 & \cellcolor{tabsecond}0.062 & \cellcolor{tabsecond}0.238 \\
\methodname  & raw & 1/1000 & \cellcolor{tabfirst}0.043 & \cellcolor{tabfirst}0.045 & \cellcolor{tabfirst}0.172 \\
\midrule
MASt3R\text{-}SfM \cite{duisterhof2025mast3r} & raw & 1/2000 & \cellcolor{tabsecond}0.124 & \cellcolor{tabsecond}0.127 & \cellcolor{tabsecond}0.380 \\
\methodname  & raw & 1/2000 & \cellcolor{tabfirst}0.089 & \cellcolor{tabfirst}0.068 & \cellcolor{tabfirst}0.244 \\
\midrule
MASt3R\text{-}SfM \cite{duisterhof2025mast3r} & raw & 1/4000 & \cellcolor{tabsecond}2.484 & \cellcolor{tabsecond}1.174 & \cellcolor{tabsecond}6.278\\
\methodname  & raw & 1/4000 & \cellcolor{tabfirst}0.227 & \cellcolor{tabfirst}0.178 & \cellcolor{tabfirst}0.575 \\
\bottomrule
\end{tabular}

}
\end{table}

We test the generalization capabilities of \methodname by capturing a dataset of 315 exposure-bracketed multi-view raw images for a single scene using an iPhone 16 (i.e., 9 different exposure settings with 35 images each).
We compare our approach to MASt3R-SfM~\cite{duisterhof2025mast3r} in Table~\ref{tab:supp-iphone} and find that our approach outperforms this baseline, especially at low signal-to-noise ratios corresponding to image captures with a short exposure time.
We show example images from the iPhone dataset in Fig.~\ref{fig:supp-iphone}, and we will publicly release this dataset.

\paragraph{Evaluation on clean images.}

\begin{table}
\captionof{table}{
\textbf{Clean image performance.} We compare the camera pose error of \methodname to MASt3R\text{-}SfM \cite{duisterhof2025mast3r}. 
We evaluate on all 330--500 sRGB images from each scene at the longest exposure in our captured dataset. 
We omit the depth metrics shown in Table~1 because they used the M\text{-}SfM output on clean images as the reference.
}

\label{tab:reb_pose_clean}
\vspace{-0.5em}
\centering
\setlength{\tabcolsep}{2pt}
%\resizebox{\columnwidth}{!}{% 
% \begin{tabular}{l c c c}
\small
\begin{tabular}{@{}lccc@{}}
\toprule
\textbf{Method} & \textbf{ATE}\,$\downarrow$ & \textbf{RPE T}\,$\downarrow$ & \textbf{RPE R}\,$\downarrow$ \\
\midrule
M-SfM [16] &  \cellcolor{tabfirst}0.088 & \cellcolor{tabfirst}0.006 & \cellcolor{tabsecond}0.031 \\
\methodname  & \cellcolor{tabsecond}0.104 & \cellcolor{tabfirst}0.006 & \cellcolor{tabfirst}0.030 \\
\bottomrule
\end{tabular}
%}
\vspace{-0.8em}
\end{table}

%%% BEGIN SIDE TABLE 

We provide an additional comparison in Table \ref{tab:reb_pose_clean} showing that \methodname has comparable pose estimation performance to  MASt3R\text{-}SfM \cite{duisterhof2025mast3r} on clean images, and hence generalizes to a broad SNR range.

\paragraph{Generalization across intrinsics.}
\begin{table}
\captionof{table}{
\textbf{Known vs.\ unknown intrinsics.} Average camera pose error and 3D reconstruction metrics with unknown and known intrinsics evaluated on the captured dataset with 120 images per scene. 
} 
\label{tab:reb_pose_unknown_intrinsics}
\vspace{-0.5em}
\centering
\setlength{\tabcolsep}{2pt}
\resizebox{\columnwidth}{!}{% 
\begin{tabular}{l l c c c c c}
\toprule
\multicolumn{2}{c}{} &
\multicolumn{3}{c}{\textbf{Camera pose error}} &
\multicolumn{2}{c}{\textbf{3D reconstruction}} \\
\cmidrule(lr){3-5}\cmidrule(lr){6-7}
\textbf{Method} & \textbf{Intrinsics} & \textbf{ATE}\,$\downarrow$ & \textbf{RPE T}\,$\downarrow$ & \textbf{RPE R}\,$\downarrow$ & \textbf{AbsRel}\,$\downarrow$ & \textbf{$\delta < 1.25$}\,$\uparrow$ \\

\midrule
% MASt3R\text{-}SfM & unknown & \cellcolor{tabthird}0.076 & 0.039 & 0.207 & \cellcolor{tabthird}0.180 & \cellcolor{tabthird}81.60 \\
\methodname & unknown & \cellcolor{tabfirst}0.043 & \cellcolor{tabfirst}0.019 & \cellcolor{tabfirst}0.121 & \cellcolor{tabfirst}0.090 & \cellcolor{tabfirst}93.63 \\
% MASt3R\text{-}SfM & known & 0.078 & \cellcolor{tabthird}0.037 & \cellcolor{tabthird}0.199 & 0.193 & 80.29 \\
\methodname & known & \cellcolor{tabsecond}0.050 & \cellcolor{tabsecond}0.020 & \cellcolor{tabfirst}0.121 & \cellcolor{tabsecond}0.091 & \cellcolor{tabsecond}93.14 \\
\bottomrule

\end{tabular}
\vspace{-1em}
}
\end{table}

Table~\ref{tab:reb_pose_unknown_intrinsics} shows that  performance of running \methodname without camera intrinsics is comparable to performance when providing intrinsics as input.

% We conduct this comparison by either providing or omitting the camera intrinsics to \methodname.
% We conduct this comparison by running \methodname with and without the camera intrinsics.
% We conduct this comparison by either providing or omitting the camera intrinsics to the fine-tuned MASt3R-SfM model.
% Further, in Table~\ref{tab:supp-iphone}, we show that the same \methodname model (without any additional fine-tuning) generalizes to images captured with a different camera (iPhone 16), also without providing the intrinsics as input. 
% \section{Rationale}
% \label{sec:rationale}
% % 
% Having the supplementary compiled together with the main paper means that:
% % 
% \begin{itemize}
% \item The supplementary can back-reference sections of the main paper, for example, we can refer to \cref{sec:intro};
% \item The main paper can forward reference sub-sections within the supplementary explicitly (e.g. referring to a particular experiment); 
% \item When submitted to arXiv, the supplementary will already included at the end of the paper.
% \end{itemize}
% % 
% To split the supplementary pages from the main paper, you can use \href{https://support.apple.com/en-ca/guide/preview/prvw11793/mac#:~:text=Delete%20a%20page%20from%20a,or%20choose%20Edit%20%3E%20Delete).}{Preview (on macOS)}, \href{https://www.adobe.com/acrobat/how-to/delete-pages-from-pdf.html#:~:text=Choose%20%E2%80%9CTools%E2%80%9D%20%3E%20%E2%80%9COrganize,or%20pages%20from%20the%20file.}{Adobe Acrobat} (on all OSs), as well as \href{https://superuser.com/questions/517986/is-it-possible-to-delete-some-pages-of-a-pdf-document}{command line tools}.

\end{document}